\documentclass[twoside,11pt]{article}
\pdfoutput=1
\usepackage{jmlr2e}
\usepackage{amsmath,subfig,float,multirow,amscd,amssymb}
\usepackage{colortbl}
\usepackage{slashbox,lscape}

\def\prodd{\displaystyle\prod}
\def\sumd{\displaystyle\sum}
\def\intd{\displaystyle\int}

\makeatletter
\def\myVCENTER#1{\vcenter{\hbox{$\m@th#1$}}}
\makeatother
\def\mathCHOICEtyped#1#2#3#4#5{#1{\mathchoice
{\myVCENTER{#2}}%
{\myVCENTER{#3}}%
{\myVCENTER{#4}}%
{\myVCENTER{#5}}}}
\def\smallerVERSION#1#2#3{\mathCHOICEtyped#1%
{\scriptstyle #2}%
{\scriptstyle #2}%
{\scriptscriptstyle #2}%
{\scriptscriptstyle #3}}
\def\puntaco{\smallerVERSION\mathbin\bullet\centerdot}

\def\bb{\mathbf{b}}

\def\Db{\mathbf{D}}

\def\Zb{\mathbf{Z}}

\def\zn{\mathbf{z}_{n\puntaco}}

\def\Ib{\mathbf{I}}

\def\Xb{\mathbf{X}}
\def\xd{\mathbf{x}_{\puntaco d}}

\def\xn{\mathbf{x}_{n\puntaco}}

\def\vn{\mathbf{v}_n}
\def\vb{\mathbf{v}}
\def\ub{\mathbf{u}}

\def\Mb{\mathbf{M}}

\def\bd{\boldsymbol{\beta}^d}

\def\Bb{\mathbf{B}}

\def\br{\mathbf{b}_{\puntaco r}}
\def\brprime{\mathbf{b}_{\puntaco r'}}

\def\rd{\boldsymbol{\rho}^d}

\def\pind{\boldsymbol{\pi}_{nd}}
\def\pindest{\boldsymbol{\widehat{\pi}}_{nd}}

\def\diag{\mathop{\rm diag}\nolimits}

\def\s2B{\sigma_B^2}

\def\MAP{{\rm MAP}}
\def\simboloX{\textrm{x}}

\def\trace{\mathop{\rm trace}\nolimits}
\def\Ocal{\mathcal{O}}

\def\Hcal{\mathcal{H}}
\def\bk{\mathbf{b}_{k \puntaco }}
\def\skrd{{(\sigma^d_{kr})}^2}
\def\b0r{{b}_{0 r}}

\jmlrheading{v}{2012}{pp-pp}{mm/12}{mm/yy}{Francisco J. R. Ruiz, Isabel Valera, Carlos Blanco and Fernando Perez-Cruz}

\ShortHeadings{Bayesian nonparametric comorbidity analysis of psychiatric disorders}{Ruiz, Valera, Blanco and Perez-Cruz}
\firstpageno{1}

\begin{document}

\title{Bayesian nonparametric comorbidity analysis of psychiatric disorders}

\author{\name Francisco J. R. Ruiz \email franrruiz@tsc.uc3m.es \\
       \addr Department of Signal Processing and Communications\\
       University Carlos III in Madrid, Spain\\
       \AND
       \name Isabel Valera \email ivalera@tsc.uc3m.es \\
       \addr Department of Signal Processing and Communications\\
       University Carlos III in Madrid, Spain\\
       \AND
       \name Carlos Blanco \email cblanco@nyspi.columbia.edu \\
       \addr Department of Psychiatry, New York State Psychiatric Institute\\Columbia University\\
       \AND
       \name Fernando Perez-Cruz \email fernando@tsc.uc3m.es \\
       \addr Department of Signal Processing and Communications\\
       University Carlos III in Madrid, Spain}
       
\editor{EditorName}

\maketitle

\begin{abstract}
The analysis of comorbidity is an open and complex research field in the branch of psychiatry, where clinical experience and several studies suggest that the relation among the psychiatric disorders may have etiological and treatment implications. In this paper, we are interested in applying latent feature modeling to find the latent structure behind the psychiatric disorders that can help to examine and explain the relationships among them. To this end, we use the large amount of information collected in the National Epidemiologic Survey on Alcohol and Related Conditions (NESARC) database and propose to model these data using a nonparametric latent model based on the Indian Buffet Process (IBP). Due to the discrete nature of the data, we first need to adapt the observation model for discrete random variables. We propose a generative model in which the observations are drawn from a multinomial-logit distribution given the IBP matrix. The implementation of an efficient Gibbs sampler is accomplished using the Laplace approximation, which allows integrating out the weighting factors of the multinomial-logit likelihood model. We also provide a variational inference algorithm for this model, which provides a complementary (and less expensive in terms of computational complexity) alternative to the Gibbs sampler allowing us to deal with a larger number of data.  Finally, we use the model  to analyze comorbidity among the psychiatric disorders diagnosed by experts from the NESARC database.
\end{abstract}


\section{Introduction}\label{sec:Introduction}

Health care increasingly needs to address the management of individuals with multiple coexisting diseases, who are now the norm, rather than the exception. In the United States, about $80\%$ of Medicare spending is devoted to patients with four or more chronic conditions, with costs growing as the number of chronic conditions increases \cite[]{Wolff2002}. This explains the growing interest of researchers in the impact of comorbidity on a range of outcomes, such as mortality, health-related quality of life, functioning, and quality of health care. However, attempts to study the impact of comorbidity are complicated by the lack of consensus about how to define and measure it \cite[]{ValderasComorbidity}.

Comorbidity becomes particularly relevant in psychiatry, where clinical experience and several studies suggest that the relation among the psychiatric disorders may have etiological and treatment implications. Several studies have focused on the search of the underlying interrelationships among psychiatric disorders, which can be useful to analyze the structure of the diagnostic classification system, and guide treatment approaches for each disorder \cite[]{Blanco2012}. \cite{Krueger1999} found that $10$ psychiatric disorders (available in the National Comorbidity Survey) can be explained by only two correlated factors, one corresponding to internalizing disorders and the other to externalizing disorders. The existence of the internalizing and the externalizing factors was also confirmed by \cite{Kotov2011}. More recently, factor analysis has been used to find the latent feature structure under $20$ common psychiatric disorders, drawing on data from the National Epidemiologic Survey on Alcohol and Related Conditions (NESARC), in \cite[]{Blanco2012}. In particular, the authors found that three correlated factors, one related to externalizing, and the other two to internalizing disorders, characterized well the underlying structure of these $20$ diagnoses. From a statistical point of view, the main limitation of this study lies on the use of factor analysis, which assumes that the number of factors is known and that the observations are Gaussian distributed. However, the latter assumption does not fit the observed data, since they are discrete in nature.

In order to avoid the model selection step needed to infer the number of factors in factor analysis, we can resort to Bayesian nonparametric tools, which allow an open-ended number of degrees of freedom in a model \cite[]{Jordan10}. In this paper, we apply the Indian Buffet Process (IBP) \cite[]{IBP}, because it allows us to infer which latent features influence the observations and how many features there are. We adapt the observation model for discrete random variables, as the discrete nature of the data does not allow using the standard Gaussian observation model. There are several options for modeling discrete outputs given the hidden latent features, like a Dirichlet distribution or sampling from the features, but we opted for the generative model partially introduced by \cite{SuicidasNIPS}, in which the observations are drawn from a multinomial-logit distribution, because it resembles the standard Gaussian observation model, as the observation probability distribution depends on the IBP matrix weighted by some factors. 

The IBP model combined with discrete observations has already been tackled in several related works. \cite{Williamson} propose a model that combines properties from both the hierarchical Dirichlet process (HDP) and the IBP, called IBP compound Dirichlet (ICD) process. They apply the ICD to focused topic modeling, where the instances are documents and the observations are words from a finite vocabulary, and focus on decoupling the prevalence of a topic in a document and its prevalence in all documents. 
Despite the discrete nature of the observations under this model, these assumptions are not appropriate for observations such as the set of possible diagnoses or responses to the questions from the NESARC database, since categorical observations can only take values from a finite set where elements do not present any particular ordering. \cite{Titsias} introduced the infinite gamma-Poisson process as a prior probability distribution over non-negative integer valued matrices with a potentially infinite number of columns, and he applied it to topic modeling of images. In this model, each (discrete) component in the observation vector of an instance depends only on one of the active latent features of that object, randomly drawn from a multinomial distribution. Therefore, different components of the observation vector might be equally distributed. Our model is more flexible in the sense that it allows different probability distributions for every component in the observation vector, which is accomplished by weighting differently the latent variables. Furthermore, a preliminary version of this model has been successfully applied to identify the factors that model the risk of suicide attempts \cite[]{SuicidasNIPS}.

The rest of the paper is organized as follows. In Section~\ref{sec:IBP}, we review the IBP model and the basic Gibbs sampling inference for the IBP, and set the notation used throughout the paper. In Section~\ref{sec:observation}, we propose the generative model which combines the IBP with discrete observations generated from a multinomial-logit distribution. In this section, we focus on the inference based on the Gibbs sampler, where we make use of the Laplace approximation to integrate out the random weighting factors in the observation model. In Section~\ref{sec:variational}, we develop a variational inference algorithm that presents lower computational complexity than the Gibbs sampler. In Section \ref{sec:experiments}, we validate our model on synthetic data and apply it over the real data extracted from the NESARC database. Finally, Section~\ref{sec:conclusions} is devoted to the conclusions.

\section{The Indian Buffet Process}\label{sec:IBP}

Unsupervised learning aims to recover the latent structure responsible for generating the observed properties of a set of objects. In latent feature modeling, the properties of each object can be represented by an unobservable vector of latent features, and the observations are generated from a distribution determined by those latent feature values. Typically, we have access to the set of observations and the main goal of latent feature modeling is to find out the latent variables that represent the data.

The most common nonparametric tool for latent feature modeling is the Indian Buffet Process (IBP). The IBP places a prior distribution over binary matrices, in which the number of rows is finite but the number of columns (features) $K$ is potentially unbounded, i.e., $K \rightarrow \infty$. This distribution is invariant to the ordering of the features and can be derived by taking the limit of a properly defined distribution over $N \times K$ binary matrices as $K$ tends to infinity \cite[]{IBP}, similarly to the derivation of the Chinese restaurant process as the limit of a Dirichlet-multinomial model \cite[]{Aldous1985}. However, given a finite number of data points $N$, it ensures that the number of non-zero columns, namely, $K_+$, is finite with probability one.

\begin{figure}[th]
\centering
\includegraphics[width=0.53\textwidth]{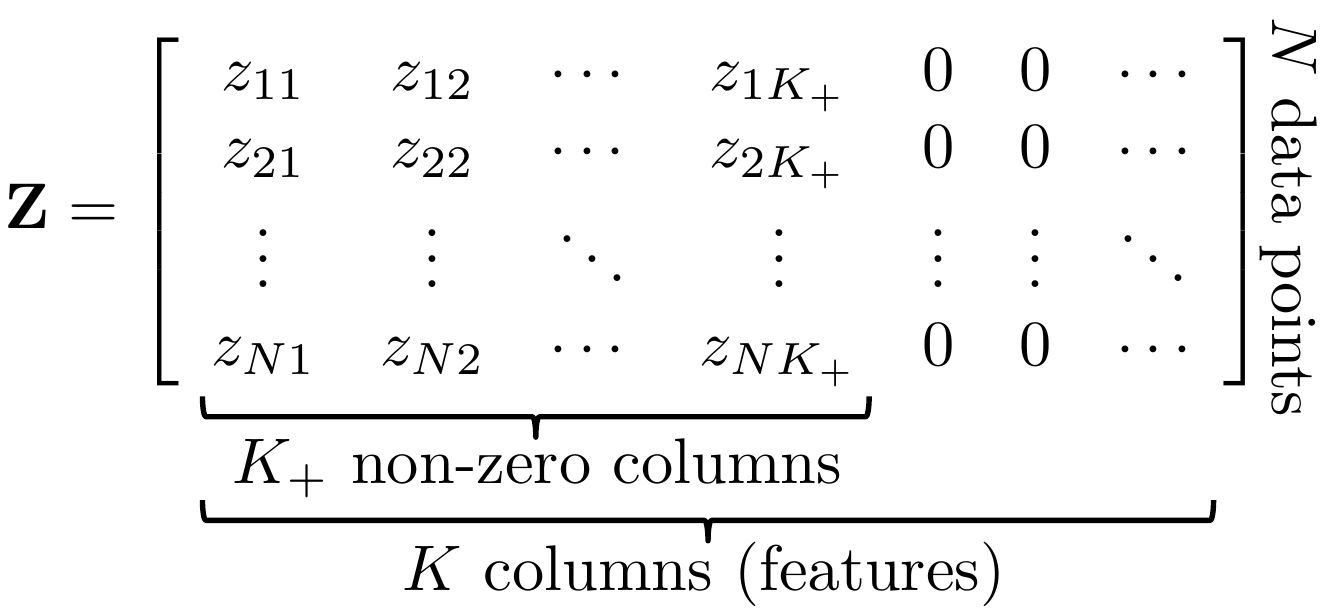}
\caption{Illustration of an IBP matrix.}
\label{fig:IBP_esquemaZ}
\end{figure}

Let $\Zb$ be a random $N\times K$ binary  matrix distributed following an IBP, i.e.,  $\Zb \sim \mathrm{IBP}(\alpha)$, where $\alpha$ is the concentration parameter of the process, which controls the number of non-zero columns $K_+$. The $n^{th}$ row of $\Zb$, denoted by $\zn$, represents the vector of latent features of the $n^{th}$ data point, and every entry $nk$ is denoted by $z_{nk}$. Note that each element $z_{nk} \in \{0,1\}$ indicates whether the $k^{th}$ feature contributes to the $n^{th}$ data point. Since only the $K_+$ non-zero columns of $\Zb$ contain the features of interest, and due to the exchangeability property of the features under the IBP prior, they are usually grouped in the left hand side of the matrix, as illustrated in Figure~\ref{fig:IBP_esquemaZ}.

Given a binary latent feature matrix $\Zb$, we assume that the $N \times D$ observation matrix $\Xb$, where the $n^{th}$ row contains a $D$-dimensional observation vector $\xn$, is distributed according to a probability distribution $p(\Xb | \Zb)$. For instance, in the standard observation model by \cite{IBP}, $p(\Xb | \Zb)$ is a Gaussian probability density function. Throughout the paper, we denote by $\xd$ the $d^{th}$ column of $\Xb$, and the elements in $\Xb$ by $x_{nd}$.

\subsection{The stick-breaking construction}\label{subsec:StickBreaking}

The stick-breaking construction of the IBP is an equivalent representation of the IBP prior, useful for inference algorithms other than Gibbs sampling, such as slice sampling or variational inference algorithms \cite[]{StickBreakingIBP,DoshiVelez}.

In this representation, the probability of each latent feature being active is represented explicitly by a random variable. In particular, the probability of feature $z_{nk}$ taking value $1$ is denoted by $\omega_k$, i.e.,
\begin{equation}
z_{nk}\sim \textrm{Bernouilli}(\omega_k).
\end{equation}
Since this probability does not depend on $n$, the stick-breaking representation explicitly shows that the ordering of the data does not affect the distribution.

The probabilities $\omega_k$ are, in turn, generated by first drawing a sequence of independent random variables $v_1,v_2,\ldots$ from a beta distribution of the form
\begin{equation}
v_{k}\sim \textrm{Beta}(\alpha,1).
\end{equation}
Given the sequence of variables $v_1,v_2,\ldots$, the probability $\omega_1$ is assigned to $v_1$, and each subsequent $\omega_k$ is obtained as
\begin{equation}
\omega_{k} = \prodd_{i=1}^{k}v_i,
\end{equation}
resulting in a decreasing sequence of probabilities $\omega_k$. Specifically, the expected probability of feature $z_{nk}$ being active decreases exponentially with the index $k$.

\begin{figure}[th]
\centering
\includegraphics[width=0.4\textwidth]{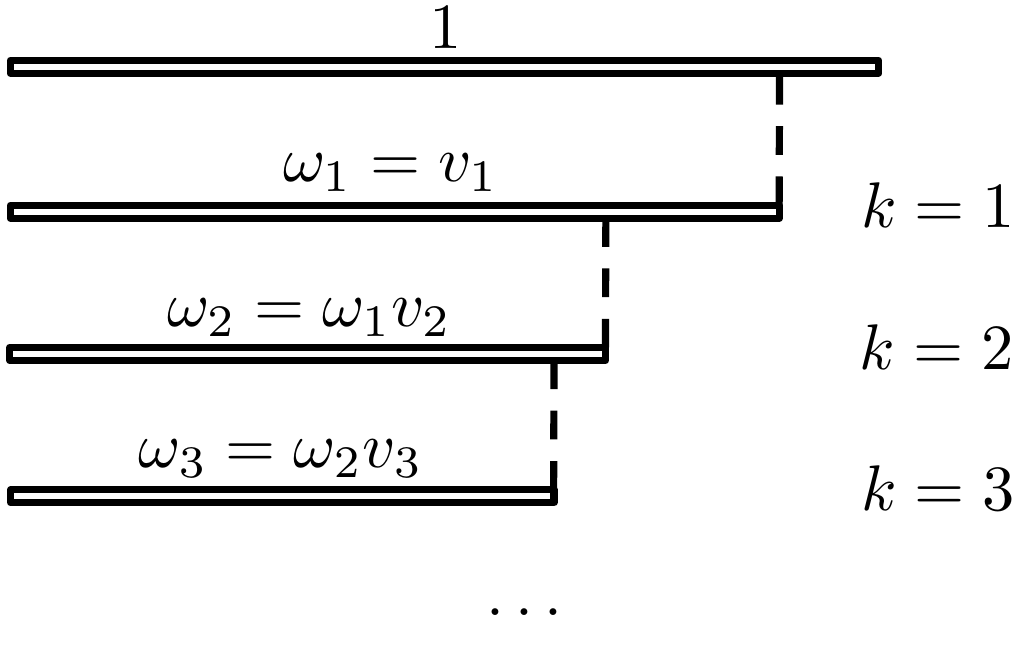}
\caption{Illustration of the stick-breaking construction of the IBP.}
\label{fig:IBP_stickbreaking}
\end{figure}

This construction can be understood with the stick-breaking process illustrated in Figure~\ref{fig:IBP_stickbreaking}. Starting with a stick of length $1$, at each iteration $k = 1,2,\ldots$, a piece is broken off at a point $v_k$ relative to the current length of the stick. The variable $\omega_k$ corresponds to the length of the stick just broken off, and the other piece of the stick is discarded.

\subsection{Inference}\label{subsec:InferenceIBP}

Markov Chain Monte Carlo (MCMC) methods have been broadly applied to infer the latent structure $\Zb$ from a given observation matrix $\Xb$ (see, e.g., \cite[]{IBP, Williamson, IFHMM, Titsias}), being Gibbs sampling the standard method of choice. This algorithm iteratively samples the value of each element $z_{nk}$ given the remaining variables, i.e., it samples from
\begin{equation}\label{eq:p_z}
 p(z_{nk}=1|\Xb,\Zb_{\neg nk} )\propto p(\Xb|\Zb)p(z_{nk}=1|\Zb_{\neg nk} ),
\end{equation}
where $\Zb_{\neg nk}$ denotes all the entries of $\Zb$ other than $z_{nk}$. The conditional distribution $p(z_{nk}=1|\Zb_{\neg nk} )$ can be readily derived from the exchangeable IBP and can be written as
\begin{equation}
 p(z_{nk}=1|\Zb_{\neg nk})=\frac{m_{-n,k}}{N},
\end{equation}
where $m_{-n,k}$ is the number of data points with feature $k$, not including $n$, i.e., $m_{-n,k}= \sum_{i \neq n} z_{ik}$. For each data point $n$, after having sampled all elements $z_{nk}$ for the $K_+$ non-zero columns in $\Zb$, the algorithm samples from a distribution (where the prior is a Poisson distribution with mean $\alpha/N$) a number of new features necessary to explain that data point.

Although MCMC methods perform exact inference, they typically suffer from high computational complexity. To solve this limitation, variational inference algorithms can be applied instead at a lower computational cost, at the expense of performing approximate inference \cite[]{IntroVariational}. A variational inference algorithm for the IBP under the standard Gaussian observation model is presented by \cite{DoshiVelez}. This algorithm makes use of the stick breaking construction of the IBP, summarized above.

\section{Observation model}\label{sec:observation}

Unlike the standard Gaussian observation model, let us consider discrete observations, i.e., each element $x_{nd}\in\{1,\ldots, R_d\}$, where this finite set contains the indexes to all the possible values of $x_{nd}$. For simplicity and without loss of generality, we consider that $R_d=R$, but the following results can be readily extended to a different cardinality per input dimension, as well as mixing continuous variables with discrete variables, since given the latent feature matrix $\Zb$ the columns of $\Xb$ are assumed to be independent.

We introduce the $K \times R$ matrices $\Bb^d$ and the length-$R$ row vectors $\bb_0^d$ to model the probability distribution over $\Xb$, such that $\Bb^d$ links the latent features with the $d^{th}$ column of the observation matrix $\Xb$, denoted by $\xd$, and $\bb_0^d$ is included to model the bias term in the distribution over the data points. This bias term plays the role of a latent variable that is always active. For a categorical observation space, if we do not have a bias term and all latent variables are inactive, the model assumes that all the outcomes are independent and equally likely, which is not a suitable assumption in most cases. In our application, the bias term is used to model the people that do not suffer from any disorder and it captures the baseline diagnosis in the general population. Additionally, this bias term simplifies the inference since the latent features of those subjects that are not diagnosed any disorder do not need to be sampled. 

\begin{figure}[th]
\centering
\includegraphics[width=0.3\textwidth]{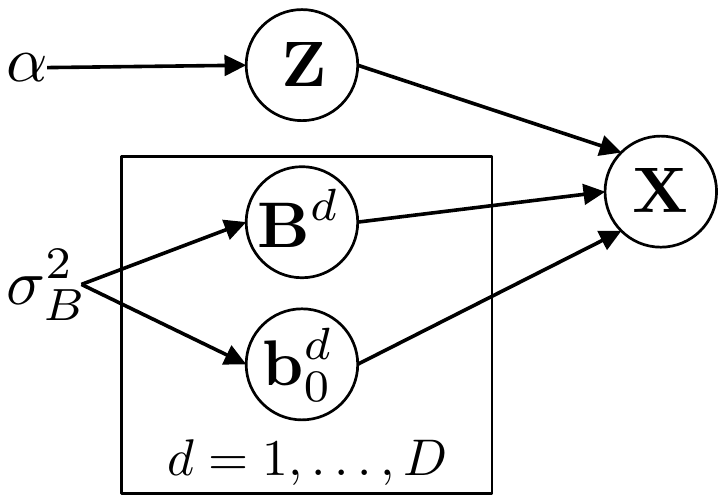}
\caption{Graphical probabilistic model of the IBP with discrete observations.}
\label{fig:IBPdiscretoObsModel}
\end{figure}

Hence, we assume that the probability of each element $x_{nd}$ taking value $r$ ($r=1, \ldots, R$), denoted by $\pi_{nd}^{r}$, is given by the multiple-logistic function, i.e.,
\begin{equation}\label{eq:LikLogitExt}
  \pi_{nd}^{r}=p(x_{nd}=r | \zn, \Bb^d, \bb_0^d)=\frac{\displaystyle\exp{(\zn \br^d+b_{0r}^d)}}{\sumd_{r'=1}^{R}\exp{(\zn \brprime^d+b_{0r'}^d)}},
\end{equation}
where $\br^d$ denotes the $r^{th}$ column of $\Bb^d$ and $b_{0r}^d$ denotes the $r^{th}$ element of vector $\bb_0^d$. Note that the matrices $\Bb^d$ are used to weight differently the contribution of each latent feature to every component $d$, similarly as in the standard Gaussian observation model in \cite[]{IBP}. We assume that the mixing vectors $\br^d$ are Gaussian distributed with zero mean and covariance matrix $\boldsymbol{\Sigma}_b = \s2B \Ib$, and the elements $b_{0r}^d$ are also Gaussian distributed with zero mean and variance $\s2B$. The corresponding graphical model is shown in Figure~\ref{fig:IBPdiscretoObsModel}.

The choice of the observation model in Eq.~\ref{eq:LikLogitExt}, which combines the multiple-logistic function with Gaussian parameters, is based on the fact that it induces dependencies among the  probabilities $\pi_{nd}^{r}$ that cannot be captured with other distributions, such as the Dirichlet distribution \cite[]{Blei}. Furthermore, this multinomial-logistic normal distribution has been widely used to define probability distributions over discrete random variables (see, e.g., \cite[]{GPs, Blei}). 

We consider that elements $x_{nd}$ are independent given the latent feature matrix $\Zb$, the weighting matrices $\Bb^d$ and the weighting vectors $\bb_0^d$. Then, the likelihood for any matrix $\Xb$ can be expressed as
\begin{equation}\label{eq: PX_BZ}
	p(\Xb | \Zb,\Bb^1,\ldots,\Bb^D,\bb_0^1,\ldots,\bb_0^D) = \prodd_{n=1}^N \prodd_{d=1}^D p(x_{nd} | \zn,\Bb^d,\bb_0^d) = \prodd_{n=1}^N \prodd_{d=1}^D \pi_{nd}^{x_{nd}}.
\end{equation}

\subsection{Laplace approximation for Gibbs sampling inference}\label{sec:laplace}

In Section \ref{sec:IBP}, the (heuristic) Gibbs sampling algorithm for posterior inference over the latent variables of the IBP, detailed in \cite[]{IBP}, has been briefly reviewed. To sample from Eq.~\ref{eq:p_z}, we need to integrate out $\Bb^d$ and $\bb_0^d$ in \eqref{eq: PX_BZ}, as sequentially sampling from the posterior distribution of these variables is intractable, for which an approximation is required. We rely on the Laplace approximation to integrate out the parameters $\Bb^d$ and $\bb_0^d$ for simplicity and ease of implementation. We first consider the finite form of the proposed model, where $K$ is bounded.


We can simplify the notation in Eqs.~\ref{eq:LikLogitExt} and \ref{eq: PX_BZ} by considering an extended latent feature matrix $\Zb$ of size $N\times (K+1)$, in which the elements of the first column are equal to one, and $D$ extended weighting matrices $\Bb^d$ of size $(K+1)\times R$, in which the first row equals the vector $\bb_0^d$. With these definitions, Eq.~\ref{eq:LikLogitExt} can be rewritten as
\begin{equation}\label{eq:LikLogit}
  \pi_{nd}^{r}=p(x_{nd}=r | \zn, \Bb^d)=\frac{\displaystyle\exp{(\zn \br^d)}}{\sumd_{r'=1}^{R}\exp{(\zn \brprime^d)}}.
\end{equation}
Unless otherwise specified, we use the simplified notation throughout this section. For this reason, the index $k$ over the latent variables takes the values in $\{0,1,\ldots,K\}$, with $z_{n0}=1$ for all $n$.

Recall that our model assumes independence among the observations given the hidden latent variables. Then, the posterior $p(\Bb^1,\ldots,\Bb^D | \Xb , \Zb)$ factorizes as
\begin{equation}
	p(\Bb^1,\ldots,\Bb^D | \Xb , \Zb) =\prod_{d=1}^Dp(\Bb^d | \xd , \Zb) =\prod_{d=1}^D \frac{p(\xd | \Bb^d,\Zb)p(\Bb^d)}{p(\xd|\Zb)}.
\end{equation}
Hence, we only need to deal with each term $p(\Bb^d | \xd , \Zb)$ individually. The marginal likelihood $p(\xd|\Zb)$, which we are interested in, can be obtained as
\begin{equation}\label{eq:marginalLikInt}
 p(\xd|\Zb)=\intd p(\xd | \Bb^d,\Zb)p(\Bb^d) d\Bb^d.
\end{equation}
Although the prior $p(\Bb^d)$ is Gaussian, due to the non-conjugacy with the likelihood term, the computation of this integral, as well as the computation of the posterior $p(\Bb^d | \xd , \Zb)$, turns out to be intractable.

Following a similar procedure as in Gaussian processes for multiclass classification \cite[]{GPs}, we approximate the posterior $p(\Bb^d | \xd , \Zb)$ as a Gaussian distribution using Laplace's method. In order to obtain the parameters of the Gaussian distribution, we define $f(\Bb^d) $ as the un-normalized log-posterior of $p(\Bb^d | \xd , \Zb)$, i.e., 
\begin{equation}\label{eq:PsiDef}
\begin{split}
 f(\Bb^d) & = \log p(\xd | \Bb^d,\Zb)+\log p(\Bb^d). 
\end{split}
\end{equation}

As proven in Appendix~\ref{app:Laplace}, the function $f(\Bb^d)$ is a strictly concave function of $\Bb^d$ and therefore it has a unique maximum, which is reached at $\Bb^d_{\MAP}$, denoted by the subscript `MAP' (\textit{maximum a posteriori}) because it coincides with the mean of the Gaussian distribution in the Laplace's approximation. We resort to Newton's method to compute $\Bb^d_{\MAP}$.


We stack the columns of $\Bb^d$ into $\bd$, i.e., $\bd=\Bb^d(:)$ for avid Matlab users. The posterior $p(\Bb^d | \xd , \Zb)$ can be approximated as
\begin{equation}\label{eq:int px_z}
 p(\bd | \xd , \Zb)  \approx \mathcal{N}\left(\bd \left| \bd_{\MAP}, \left. (-\nabla\nabla f )\right|_{\bd_{\MAP}}\right.\right),
\end{equation}
where $\bd_{\MAP}$ contains all the columns of $\Bb^d_{\MAP}$ stacked into a vector and $\nabla\nabla f$ is the Hessian of $f(\bd)$.
Hence, by taking the second-order Taylor series expansion of $f(\bd)$ around its maximum, the computation of the marginal likelihood in~\eqref{eq:marginalLikInt}
results in a Gaussian integral, whose solution can be expressed as
 \begin{align}
  &\log p(\xd|\Zb) \approx  -\displaystyle\frac{1}{2\s2B} \trace\left\{(\Bb_{\MAP}^d)^\top\Bb_{\MAP}^d\right\}\nonumber \\
  &- \frac{1}{2}\log\left| \Ib_{R(K+1)}+\s2B \sumd_{n=1}^N \left(\diag(\pindest)-(\pindest)^\top \pindest\right) \otimes (\zn^\top\zn) \right| +\log p(\xd|\Bb^d_{\MAP},\Zb),\label{eq:logMargLik}
 \end{align}
where $\pindest$ is the vector $\pind = \left[ \pi^1_{nd}, \pi^2_{nd}, \ldots, \pi^R_{nd}\right]$ evaluated at $\Bb^d=\Bb^d_{\MAP}$, and $\diag(\pindest)$ is a diagonal matrix with the values of $\pindest$ as its diagonal elements.

Similarly as in \cite[]{IBP}, it is straightforward to prove that the limit of Eq.~\ref{eq:logMargLik} is well-defined if $\Zb$ has an unbounded number of columns, i.e., as $K \rightarrow \infty$. The resulting expression for the marginal likelihood $p(\xd|\Zb)$ can be readily obtained from Eq.~\ref{eq:logMargLik} by replacing $K$ by $K_{+}$, $\Zb$ by the submatrix containing only the non-zero columns of $\Zb$, and $\Bb^d_{\MAP}$ by the submatrix containing the $K_{+}$+1 corresponding rows.

\subsection{Speeding up the matrix inversion}

In this section, we propose a method that reduces the complexity of computing the inverse of the Hessian for Newton's method (as well as its determinant in~\eqref{eq:logMargLik})  from $\Ocal(R^3K_+^3+NR^2K_+^2)$ to $\Ocal(RK_+^3+NR^2K_+^2)$, effectively accelerating the inference procedure for large values of $R$. 

Let us denote with $\Zb$ the matrix that contains only the $K_+ +1$ non-zero columns of the extended full IBP matrix. The inverse of the Hessian for Newton's method, as well as its determinant in \eqref{eq:logMargLik}, can be efficiently carried out if we rearrange the inverse of $\nabla\nabla f$ as follows:
\begin{equation}\label{eq:inverse}
   (-\nabla\nabla f)^{-1} = \left( \Db - \sumd_{n=1}^N\vn\vn^\top\right)^{-1},
\end{equation}
where $\vn= (\pind)^\top\otimes \zn^\top$ and $\Db$ is a block-diagonal matrix, in which each diagonal submatrix is given by
\begin{equation}\label{eq:Dr}
 \Db_r = \displaystyle\frac{1}{\s2B}\Ib_{K_{+}+1}+\Zb^\top\diag\left(\boldsymbol{\pi}_{\puntaco d}^r\right)\Zb,
\end{equation}
with $\boldsymbol{\pi}_{\puntaco d}^r=\left[\begin{array}{ccc}\pi^{r}_{1d},\ldots,\pi^{r}_{Nd} \end{array}\right]^\top$. Since $\vn\vn^\top$ is a rank-one matrix, we can apply the Woodbury identity \cite[]{Woodbury} $N$ times to invert the matrix $-\nabla\nabla f$, similar to the RLS (Recursive Least Squares) updates \cite[]{RLS}. At each iteration $n=1,\ldots,N$, we compute
\begin{equation}\label{eq:Woodbury_D}
 (\Db^{(n)})^{-1} = \left(\Db^{(n-1)}-\vn\vn^\top\right)^{-1}=(\Db^{(n-1)})^{-1}+\frac{(\Db^{(n-1)})^{-1}\vn\vn^\top(\Db^{(n-1)})^{-1}}{1-\vn^\top(\Db^{(n-1)})^{-1}\vn}.
\end{equation}

For the first iteration, we define $\Db^{(0)}$ as the block-diagonal matrix $\Db$, whose inverse matrix involves computing the $R$ matrix inversions of size $(K_{+}+1)\times (K_{+}+1)$ of the matrices in \eqref{eq:Dr}, which can be efficiently solved applying the Matrix Inversion Lemma. After $N$ iterations of \eqref{eq:Woodbury_D}, it turns out that $(-\nabla\nabla f)^{-1} = (\Db^{(N)})^{-1}$.

In practice, there is no need to iterate over all observations, since all subjects sharing the same latent feature vector $\zn$ and observation $x_{nd}$ can be grouped together, therefore requiring (at most) $R2^{K_+}$ iterations instead of $N$. In our applications, it provides significant savings in run-time complexity, since $R2^{K_+}\ll N$.

For the determinant in \eqref{eq:logMargLik}, similar recursions can be applied using the Matrix Determinant Lemma \cite[]{matrixDetLemma}, which states that $|\Db+\vb\ub^\top|=(1+\vb^\top\Db\ub)|\Db|$, and $|\Db^{(0)}|=\prod_{r=1}^R|\Db_r|$.

\section{Variational Inference}\label{sec:variational}

Variational inference provides a complementary (and less expensive in terms of computational complexity) alternative to MCMC methods as a general source of approximation methods for inference in large-scale statistical models \cite[]{IntroVariational}. In this section, we adapt the infinite variational approach for the linear-Gaussian model with respect to a full IBP prior introduced by \cite{DoshiVelez} to the model proposed in Section~\ref{sec:observation}. This approach assumes the (truncated) stick-breaking construction for the IBP in Section~\ref{subsec:StickBreaking}, which bounds the number of columns of the IBP matrix  by a finite (but large enough) value, $K$. Then, in the truncated stick-breaking process, $\omega_k= \prod_{i=1}^{k} v_i$ for $k\leq K$ and zero otherwise.

The hyperparameters of the model are contained in the set $\Hcal=\{ \alpha, \sigma^2_B \}$ and, similarly, $\Psi=\{\Zb,\Bb^1,\ldots,\Bb^D, \bb_0^1,\ldots,\bb_0^D, v_1, \ldots, v_K\}$ denotes the set of unobserved variables in the model. Under the truncated stick-breaking construction for the IBP, the joint probability distribution over all the variables $p(\Psi,\Xb|\Hcal)$ can be factorized as 
\begin{equation}
\begin{split}
 p(\Psi,\Xb|\Hcal)= & \prodd_{k=1}^{K}  \left(p(v_k|\alpha) \prodd_{n=1}^{N} p(z_{nk}|\{v_i\}_{i=1}^{k}) \right)
\prodd_{d=1}^D \left(p(\bb_0^d|\s2B) \prodd_{k=1}^K p(\bk^d | \s2B) \right) \\
& \times \prodd_{n=1}^{N}\prodd_{d=1}^{D} p(x_{nd} | \zn, \Bb^d, \bb_0^d),
\end{split}
\end{equation}
where $\bk^d$ is the $k^{th}$ row of matrix $ \Bb^d$.

We approximate $p(\Psi|\Xb,\Hcal)$ with the variational distribution $q(\Psi)$ given by
\begin{equation}\label{eq:variationalPdf}
q(\Psi)=\prodd_{k=1}^{K}  \left(q(v_k|\tau_{k1},\tau_{k2}) \prodd_{n=1}^{N} q(z_{nk}|\nu_{nk})  \right)
\prodd_{k=0}^{K}\prodd_{r=1}^{R} \prodd_{d=1}^{D} q(b^d_{kr} | \phi^d_{kr},\skrd),
\end{equation}
where the elements of matrix $ \Bb^d$ are denoted by  $b^d_{kr}$, and 
\begin{subequations}
\begin{align}
& q(v_k|\tau_{k1},\tau_{k2})= \mathrm{Beta}(\tau_{k1},\tau_{k2}), \\
& q(b^d_{kr} | \phi^d_{kr},\skrd) = \mathcal{N}(\phi^d_{kr},\skrd), \\
& q(z_{nk}|\nu_{nk}) = \mathrm{Bernoulli}(\nu_{nk}).
\end{align}
\end{subequations}
Inference involves optimizing the variational parameters of $q(\Psi)$ to minimize the Kullback-Leibler divergence from $q(\Psi)$ to $p(\Psi|\Xb,\Hcal)$, i.e., $D_{KL}(q||p)$. This optimization is equivalent to maximizing a lower bound on the evidence $p(\Xb|\Hcal)$, since
\begin{equation}\label{eq:defCotaVariac}
\begin{split}
 \log p(\Xb|\Hcal) & = \mathbb{E}_q \left[ \log p(\Psi,\Xb|\Hcal) \right]  + H[q] +D_{KL}(q||p) \\
 & \geqslant \mathbb{E}_q \left[ \log p(\Psi,\Xb|\Hcal) \right]  + H[q],
\end{split}
\end{equation}
where $\mathbb{E}_q[\cdot]$ denotes the expectation with respect to the distribution $q(\Psi)$, $H[q]$ is the entropy of distribution $q(\Psi)$ and
\begin{equation}
\begin{split}\label{Eq:Eq(p)}
\mathbb{E}_q \left[ \log  p(\Psi,\Xb|\Hcal) \right]&=  \sumd_{k=1}^{K}  \mathbb{E}_q \left[\log p(v_k|\alpha)  \right]+  \sumd_{d=1}^{D}\sumd_{k=1}^{K}  \mathbb{E}_q \left[\log p(\bk^d | \s2B) \right] + \sumd_{d=1}^{D} \mathbb{E}_q \left[\log p(\bb_0^d | \s2B) \right]\\
&  \quad + \sumd_{k=1}^{K}\sumd_{n=1}^{N} \mathbb{E}_q \left[\log  p(z_{nk}|\{v_i\}_{i=1}^{k}) \right] +\sumd_{n=1}^{N}\sumd_{d=1}^{D} \mathbb{E}_q \left[\log p(x_{nd} | \zn, \Bb^d,\bb_0^d) \right].
\end{split}
\end{equation}
The derivation of the lower bound in \eqref{eq:defCotaVariac} is straightforward, except for the terms $\mathbb{E}_q \left[\log p(z_{nk}|\{v_i\}_{i=1}^{k}) \right]$ and $\mathbb{E}_q \left[\log p(x_{nd} | \zn, \Bb^d, \bb_0^d) \right]$ in \eqref{Eq:Eq(p)}, which have no closed-form solution, so we instead need to bound them. Deriving these bounds leads to a new bound $\mathcal{L}(\Hcal, \Hcal_q)$, which can be obtained in closed-form, such that $\log p(\Xb|\Hcal)\geq \mathcal{L}(\Hcal, \Hcal_q)$, being $\Hcal_q$ the full set of variational parameters. The final expression for $\mathcal{L}(\Hcal, \Hcal_q)$, as well as the details on the derivation of the bound, are provided in Appendix~\ref{app:lowerBound}.

In order to maximize the lower bound $\mathcal{L}(\Hcal, \Hcal_q)$, we need to optimize with respect to the value of the variational parameters. To this end, we can iteratively maximize the bound with respect to each variational parameter by taking the derivative of $\mathcal{L}(\Hcal, \Hcal_q)$ and setting it to zero. This procedure readily leads to the following fixed-point equations:
\begin{enumerate}
\item For the variational Beta distribution $q(v_k|\tau_{k1},\tau_{k2})$,
\begin{subequations}
\begin{align}
 \tau_{k1}&=\alpha + \sumd_{m=k}^{K} \left( \sumd_{n=1}^{N}  \nu_{nm}\right) + \sumd_{m=k+1}^{K} \left( N - \sumd_{n=1}^{N} \nu_{nm} \right) \left(\sumd_{i=k+1}^{m} \lambda_{mi}  \right),   \\
 \tau_{k2}&=1+ \sumd_{m=k}^{K} \left( N - \sumd_{n=1}^{N} \nu_{nm} \right)  \lambda_{mk}.
\end{align}
\end{subequations}

\item For the Bernoulli distribution $q(z_{nk}|\nu_{nk})$,
\begin{equation}
\nu_{nk}= \frac{1}{1+\exp(-A_{nk})},
\end{equation}
where
\begin{equation}
\begin{split}
A_{nk}&=\sumd_{i=1}^{k} \left[\psi(\tau_{i1})- \psi(\tau_{i1}+ \tau_{i2}) \right] -\left[\sumd_{m=1}^{k} \lambda_{km} \psi( \tau_{m2}) + \sumd_{m=1}^{k-1} \left(\sumd_{n=m+1}^{k} \lambda_{kn} \right) \psi( \tau_{m1}) \right.  \\
& \left.   \quad - \sumd_{m=1}^{k} \left( \sumd_{n=m}^{k} \lambda_{kn} \right) \psi( \tau_{m1} + \tau_{m2}) - \sumd_{m=1}^{k} \lambda_{km} \log (\lambda_{km})\right] \\
& + \sumd_{d=1}^{D} \left(\phi^d_{kx_{nd}}  - \xi_{nd} \sumd_{r=1}^{R} \Bigg[ \exp\left(\phi^d_{0r}+\frac{1}{2} (\sigma^{d}_{0r})^2 \right) \left( 1-\exp\left(\phi^d_{kr}+\frac{1}{2} \skrd \right)\right) \times \right.  \\
& \quad \quad  \quad  \left. \times \prodd_{k'\neq k} \left( 1-\nu_{nk'}+ \nu_{nk'} \exp\left(\phi^d_{k'r}+\frac{1}{2} {(\sigma^d_{k'r})}^2 \right)\right)  \Bigg]\right),
\end{split}
\end{equation}
and $\psi(\cdot)$ stands for the digamma function \cite[p.~258--259]{Abramowitz}.

\item For the feature assignments, which are Bernoulli distributed given the feature probabilities, we have lower bounded $\mathbb{E}_q \left[\log p(z_{nk}|\{v_i\}_{i=1}^{k})   \right]$ by using the multinomial approach in \cite[]{DoshiVelez} (see Appendix~\ref{app:lowerBound} for further details). This approximation introduces the auxiliary multinomial distribution $\boldsymbol{\lambda}_k= [\lambda_{k1},\ldots,\lambda_{kk}]$, where each $\lambda_{ki}$ can be updated as
\begin{equation}
\lambda_{ki} \propto \exp \left(  \psi(\tau_{i2}) + \sumd_{m=1}^{i-1}\psi(\tau_{m1}) - \sumd_{m=1}^{i} \psi(\tau_{m1}+ \tau_{m2}) \right),
\end{equation}
where the proportionality ensures that $\boldsymbol{\lambda}_k$ is a valid distribution.

\item The maximization with respect to the variational parameters $ \phi^d_{kr}$, $\phi^d_{0r}$, $\skrd$, and $(\sigma^d_{0r})^2$ has no analytical solution, and therefore, we need to resort to a numerical method to find the maximum, such as Newton's method or conjugate gradient algorithm, for which the first and the second derivatives\footnote{ Note that the second derivatives are strictly negative and, therefore, the maximum with respect to each parameter is unique.} (given in Appendix~\ref{app:derivatives}) are required.

\item Finally, we lower bound the likelihood term $\mathbb{E}_q \left[\log p(x_{nd} | \zn, \Bb^d, \bb_0^d) \right]$ by resorting to a first-order Taylor series expansion around the auxiliary variables $\xi_{nd}^{-1}$ for $n=1,\ldots,N$ and $d=1,\ldots,D$ (see Appendix~\ref{app:lowerBound} for further details), which are optimized by the expression
\begin{equation}
\xi_{nd}= \left[ \sumd_{r=1}^{R}  \exp\left(\phi^d_{0r}+\frac{1}{2} (\sigma^{d}_{0r})^2 \right)   \prodd_{k=1}^{K} \left( 1-\nu_{nk}+ \nu_{nk} \exp\left(\phi^d_{kr}+\frac{1}{2} \skrd \right) \right) \right]^{-1}.
\end{equation}

\end{enumerate}

\section{Experiments}\label{sec:experiments}


\subsection{Inference over synthetic images}

\begin{figure}[th]
\centering
\begin{tabular}{cc}
  \subfloat[]{\label{fig:basis}\includegraphics[width=0.4\textwidth]{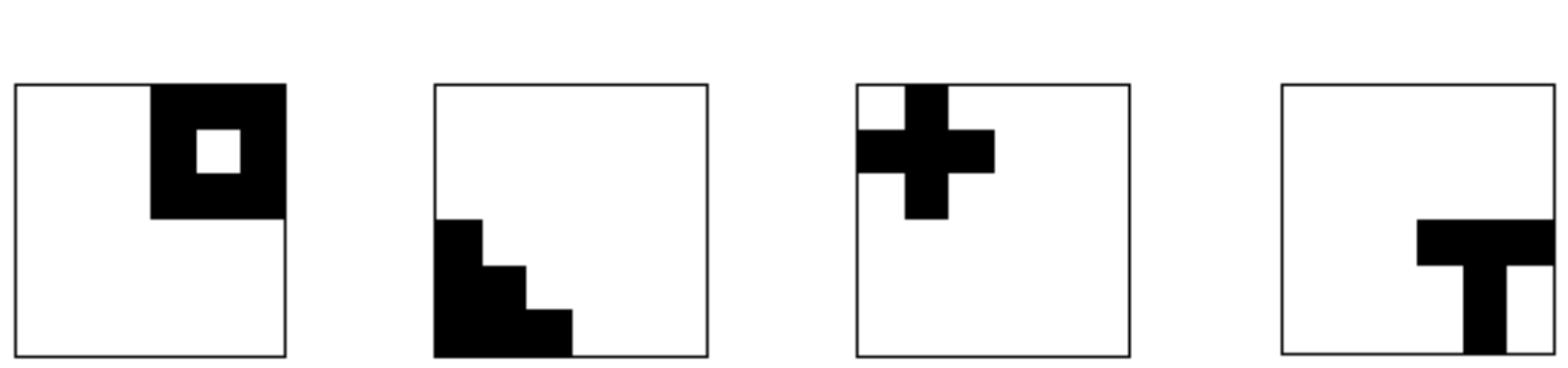}} \hspace*{2mm}
  \subfloat[]{\label{fig:latentImag}\includegraphics[width=0.4\textwidth]{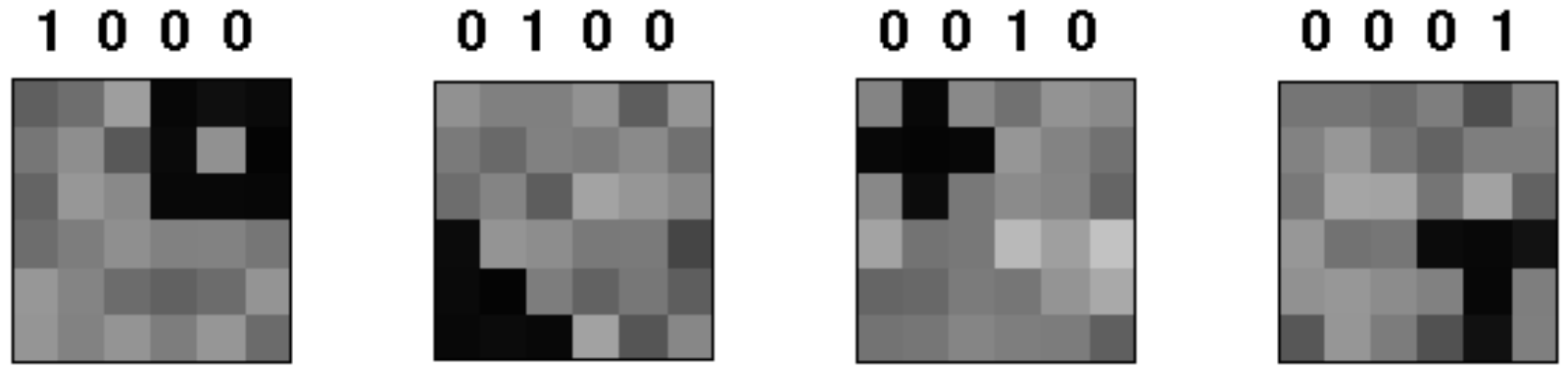}} & \multirow{2}{*}[1cm]{\includegraphics[width=0.06\textwidth]{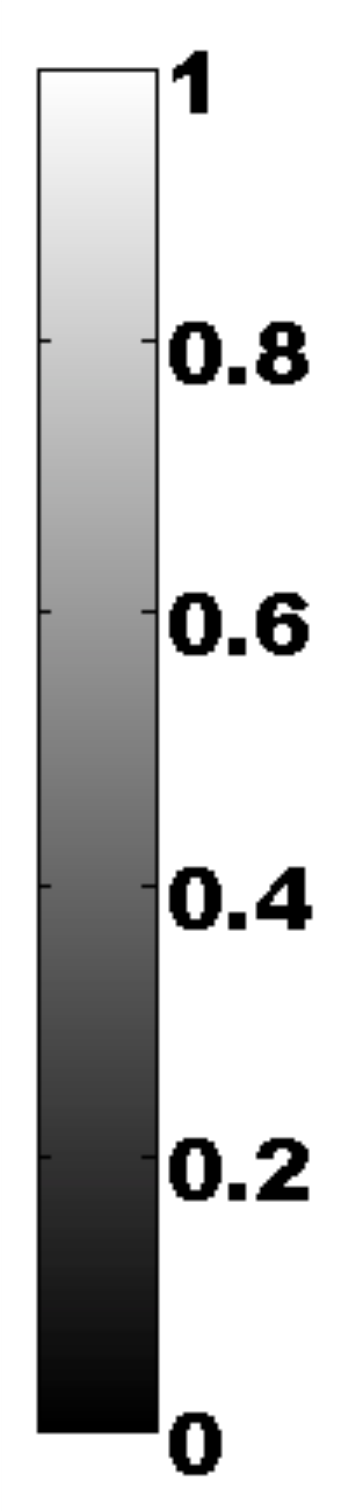}}\\
  \subfloat[ ]{\label{fig:ImSamples}\includegraphics[width=0.4\textwidth]{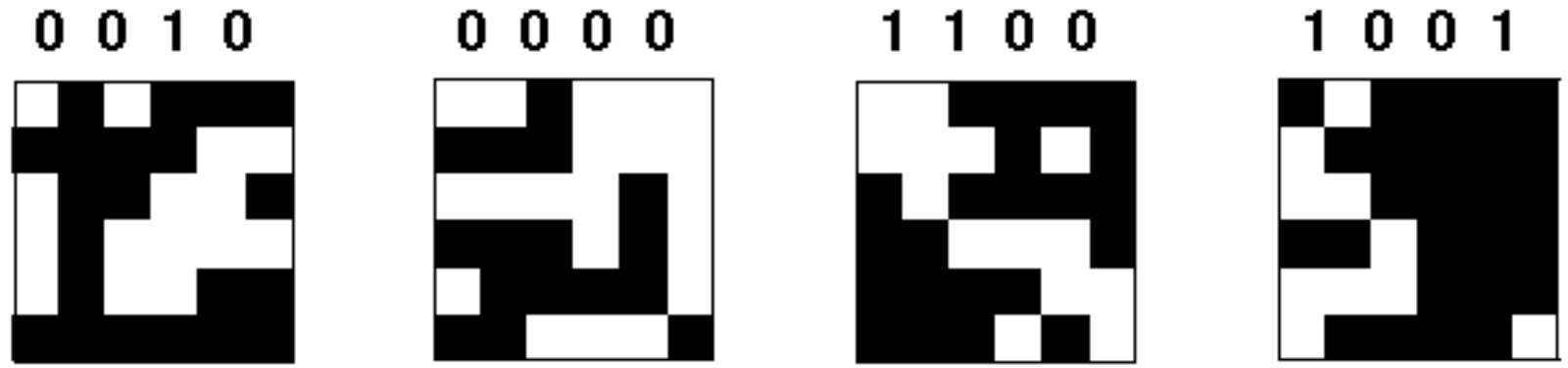}} \hspace*{2mm}
  \subfloat[ ]{\label{fig:ImProb}\includegraphics[width=0.4\textwidth]{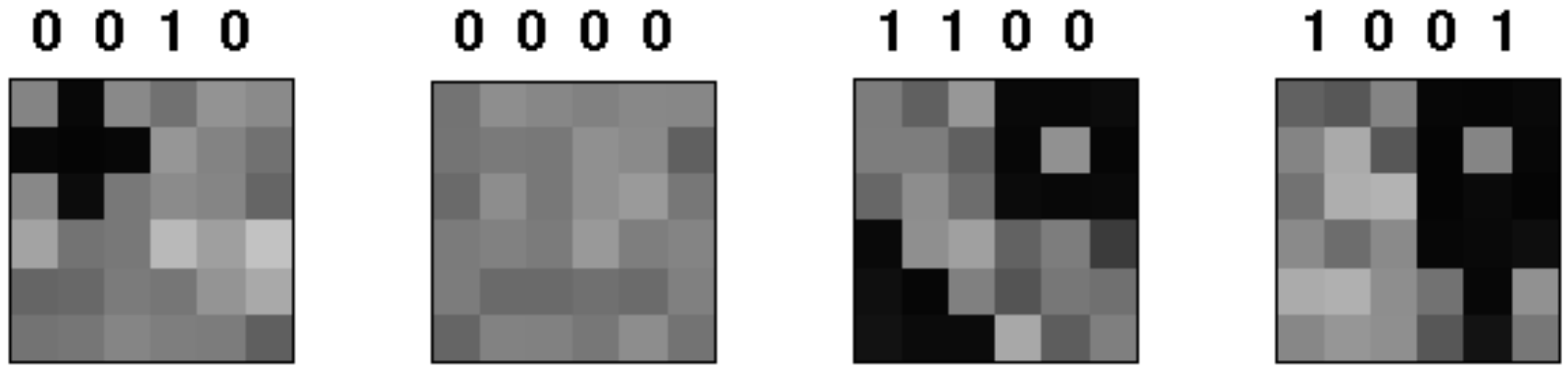}} &    \\
  \end{tabular}
  \subfloat[ ]{\label{fig:NK}\includegraphics[width=0.48\textwidth]{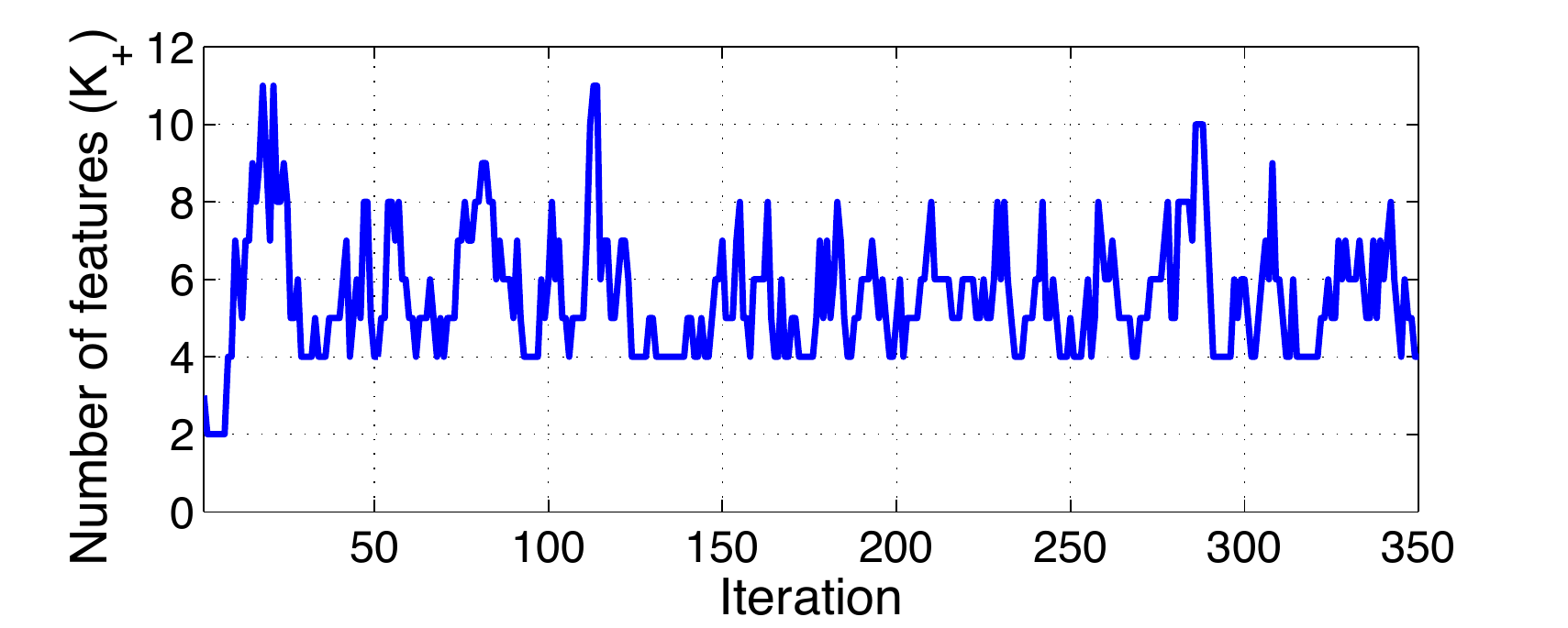}} 
  \subfloat[ ]{\label{fig:pX_Z}\includegraphics[width=0.48\textwidth]{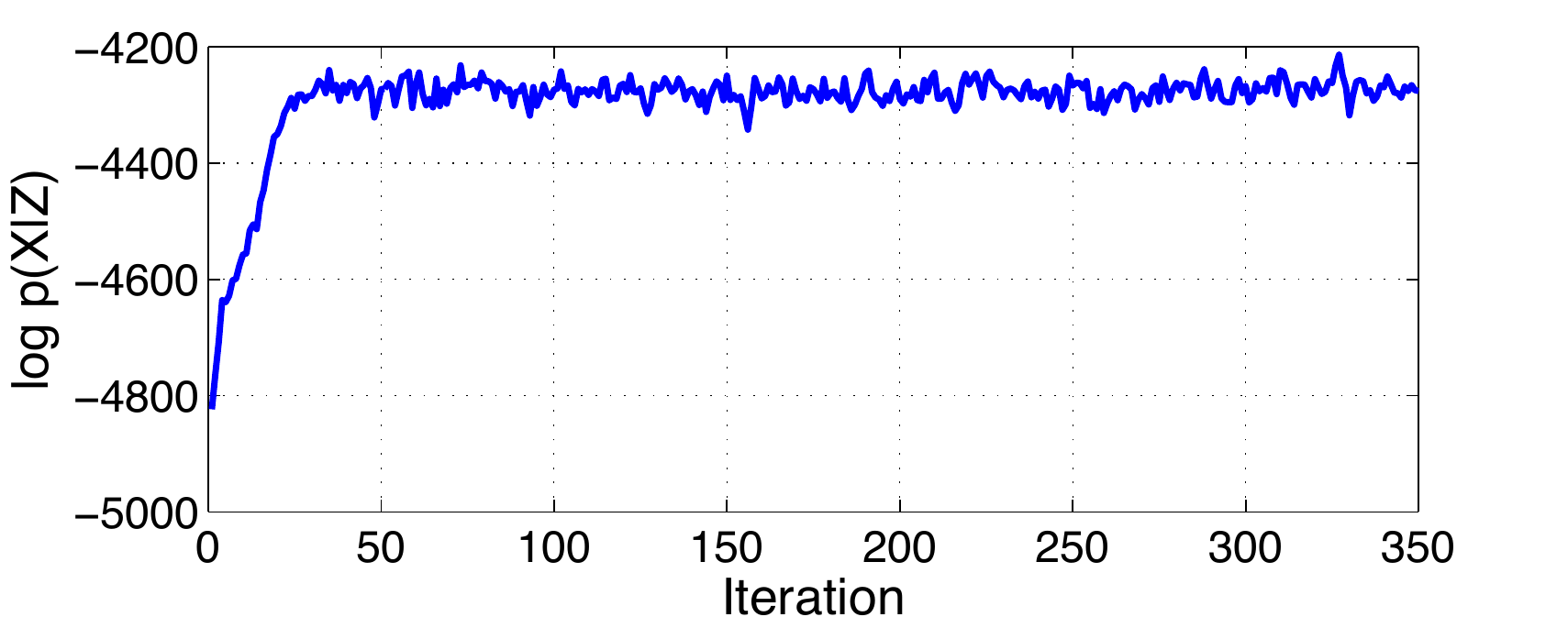}} \\ 

\caption{Experimental results of the infinite binary multinomial-logistic model over the image data set. (a) The four base images used to generate the 200 observations. (b) Probability of each pixel being white, when a single feature is active (ordered to match the images on the left), computed using the matrices $\Bb^d_{\MAP}$. (c) Four data points generated as described in the text. The numbers above each figure indicate which features are present in that image. (d) Probabilities of each pixel being white after $350$ iterations of the Gibbs sampler inferred for the four data points on (c). The numbers above each figure show the inferred value of $\zn$ for these data points. (e) The number of latent features $K_+$  and (f) the approximate log of $p(\Xb|\Zb)$ over the $200$ iterations of the Gibbs sampler.}
\label{fig:Images}
\end{figure}

We generate an illustrative example inspired by the example in \cite[]{IBP} to show that the proposed model works as expected. We define four base black-and-white images, shown in Figure~\ref{fig:basis}, that can be present with probability $0.3$, independently of the others. These base images are combined to create a binary composite image. We also multiply each white pixel independently with equiprobable binary noise, hence each white pixel in the composite image can be turned black $50\%$ of the times, while black pixels always remain black. We generate $200$ observations to learn the IBP model (several examples can be found in Figure~\ref{fig:ImSamples}). The Gibbs sampler has been initialized with $K_+=2$, setting each $z_{nk}=1$ with probability $1/2$, and setting the hyperparameters to $\alpha=0.5$ and $\s2B=1$. 

After $350$ iterations, the Gibbs sampler returns four latent features. Each of the four features recovers one of the base images with a different ordering, which is inconsequential. In Figure~\ref{fig:latentImag}, we have plotted the posterior probability for each pixel being white, when only one of the components is active. As expected, the black pixels are known to be black (almost zero probability of being white) and the white pixels have about a $50/50$ chance of being black or white, due to the multiplicative noise. The Gibbs sampler has used as many as eleven hidden features, as shown in Figure~\ref{fig:NK}, but after less than $50$ iterations, the first four features represent the base images and the others just lock on to a noise pattern, which eventually fades away.

In Figure~\ref{fig:ImProb}, we depict the posterior probability of pixels being white for the four images in Figure~\ref{fig:ImSamples}, given the inferred latent feature vectors for these observations. Note that the model behaves as expected and properly captures the generative process, even for those observations which do not possess any latent features, for which the vectors $\bb_0^d$ do not provide significant information about the black-or-white probabilities.

\subsection{Comorbidity analysis of psychiatric disorders}

In the present study, our objective is to provide an alternative to the factor analysis approach used by \cite{Blanco2012} with the IBP for discrete observations introduced in the present paper. We build an unsupervised model taking the $20$ disorders used by \cite{Blanco2012} as input data, drawn from the NESARC data.

The NESARC database was designed to estimate the prevalence of psychiatric disorders, as well as their associated features and level of disability. The NESARC had two waves of interviews (first wave in 2001-2002 and second wave in 2004-2005). For the following experimental results, we only use the data from the first wave, for which $43{,}093$ people were selected to represent the U.S. population of 18 years of age and older. Through $2{,}991$ entries, the NESARC collects data on the background of participants, alcohol and other drug use and use disorders, and other mental disorders. Public use data are currently available for this wave of data collection\footnote{See http://aspe.hhs.gov/hsp/06/catalog-ai-an-na/nesarc.htm}.

The $20$ disorders include substance use disorders (alcohol abuse and dependence, drug abuse and dependence and nicotine dependence), mood disorders (major depressive disorder (MDD), bipolar disorder and dysthymia), anxiety disorders (panic disorder, social anxiety disorder (SAD), specific phobia and generalized anxiety disorder (GAD)), pathological gambling (PG) and seven personality disorders (avoidant, dependent, obsessive-compulsive (OC), paranoid, schizoid, histrionic and antisocial personality disorders (PDs)).

We run the Gibbs sampler over $3,500$ randomly chosen subjects out of the $43{,}093$ participants in the survey, having initialized the sampler with an active feature, i.e., $K_+=1$, having set $z_{nk}=1$ randomly with probability $0.5$, and fixing $\alpha=1$ and $\s2B=1$. After convergence, we run an additional Gibbs sampler with $10$ iterations for each of the remaining subjects in the database, restricted to their latent features (i.e., we fix the latent features learned for the $3,500$ subjects to sample the feature vector of each subject). Then, we run additional iterations of the Gibbs sampler over the whole database, finally obtaining three latent features. In order to speed up the sampling procedure, we do not sample the rows of $\Zb$ corresponding to those subjects who do not suffer from any of the $20$ disorders, but instead fix these latent features to zero. The idea is that the $\bb_0^d$ terms must capture the general population that does not suffer from any psychiatric disorder, and we use the active components of the matrix $\Zb$ to characterize the disorders. 

\begin{figure}[th]
\centering
\includegraphics[width=\textwidth]{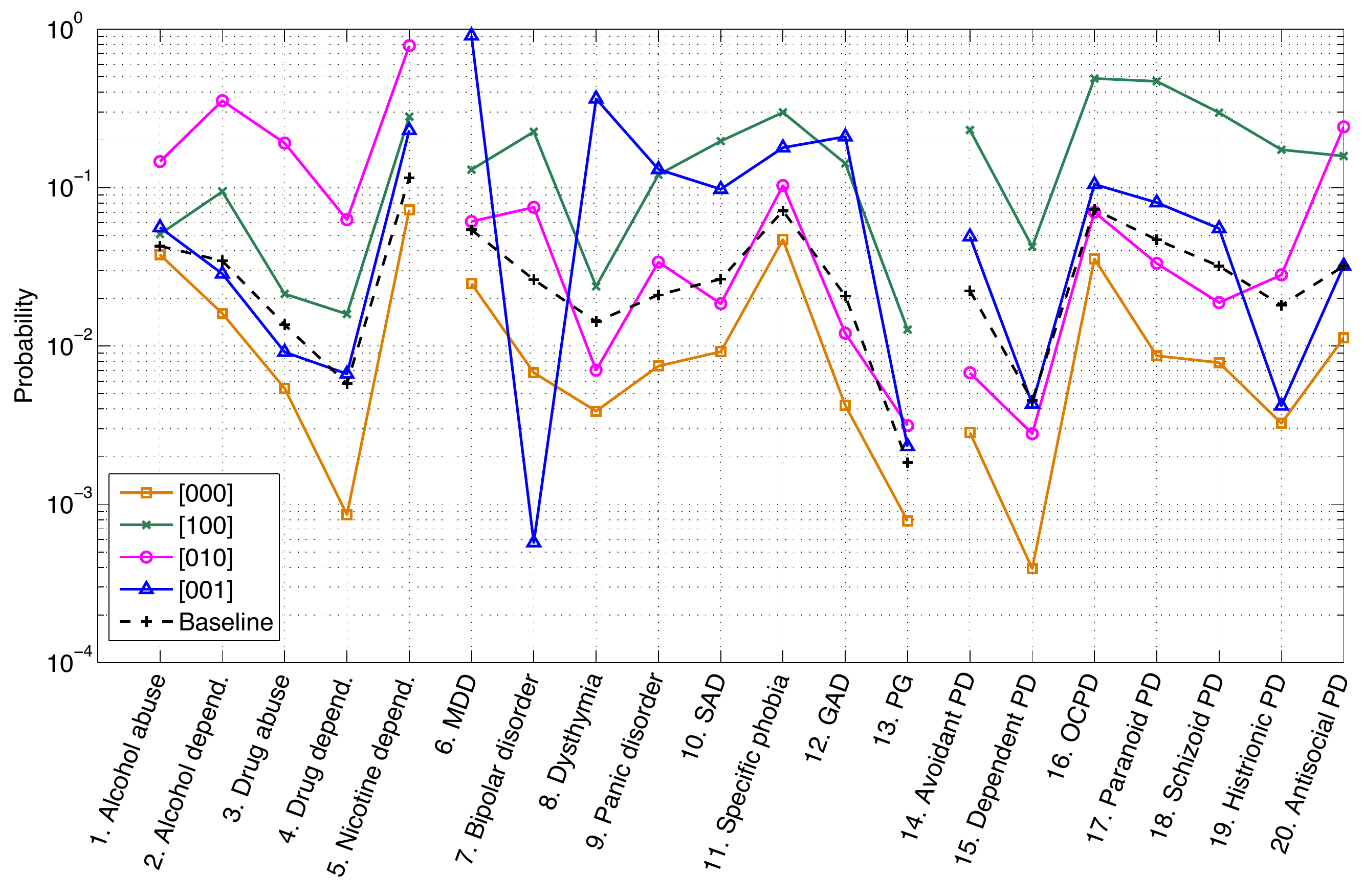}
\caption{Probabilities of suffering from the $20$ considered disorders. These probabilities have been obtained using the matrices $\Bb^d_{\MAP}$, when none or a single latent feature is active. The legend shows the latent feature vector corresponding to each curve. The baseline has been obtained taking into account the $43,093$ subjects in the database.}
\label{fig:20Q}
\end{figure}

To examine the three latent features, we plot in Figure~\ref{fig:20Q} the posterior probability of having each of the considered disorders, when none or one of the latent features is active. As expected, for those subjects who do not possess any feature, the probability of having any of the disorders is below the baseline level (due to the contribution of the vectors $\bb_0^d$), defined as the empirical probability in the full sample, i.e., taking into account the $43,093$ participants. Feature 1 increases the probability of having all the disorders, and thus seems to represent a general psychopathology factor, although it may particularly increase the risk of personality disorders. Feature 2 models substance use disorders and antisocial personality disorder, consistent with the externalizing factor identified in previous studies of the structure of psychiatric disorders \cite[]{Krueger1999, Kendler2003, Vollebergh2001, Blanco2012}. Feature 3 models mood or anxiety disorders, and thus seems to represent the internalizing factor also identified in previous studies.  

Thus, in accord to previous results from the studies on the latent structure of the comorbidity of psychiatric disorders, detailed in Section 1, we find that the patterns of comorbidity of common psychiatric disorders can be well described by a small number of latent features. In addition, nosologically related disorders, such as social anxiety disorder and avoidant personality disorder, tend to be modeled by similar features. As found in previous results \cite[]{Blanco2012}, no disorder is perfectly aligned along one single latent feature, therefore suggesting that disorders can develop through multiple etiological paths. For instance, the risk of nicotine dependence may be particularly high in individuals with a propensity towards externalization or internalization.

In Table~\ref{tab:indepFeat20Q_a}, we first show the empirical probability of possessing each latent feature, i.e., the number of subjects in the database that possess each latent feature divided by the total number of subjects. We also show in Table~\ref{tab:indepFeat20Q_b} the probability of possessing at least two features as the product of the probabilities in Table~\ref{tab:indepFeat20Q_a} (Product Probability), and also the empirical probability. We include Table~\ref{tab:indepFeat20Q} to show that the three features are nearly independent of one another, since the probability of possessing any two particular features is close to the product of the probabilities of possessing them individually. The differences in Table~\ref{tab:indepFeat20Q_b} are not statistically significant. Then, besides explicitly capturing the probability of each disorder, our model also provides a way to measure independence among the latent features. Note that although the proposed model assumes that the latent features are independent \textit{a priori}, we could have found that the empirical probability does not correspond to the product one. Therefore, the independence among the three latent features follows the model's assumption and, from a psychiatric perspective, it also shows that the three factors (internalizing, externalizing and general psychopathology factor) are independent one another, i.e., suffering from one group of disorders does not imply an increased probability of suffering from any other group of disorders.

Finally, we remark that we have also applied the variational inference algorithm to study the  comorbidity patterns of psychiatric disorders but, since both algorithms (the variational and the Gibbs sampler) infer the same three latent features, we only plot the results for the Gibbs sampling algorithm in this section and apply the variational inference algorithm in next section.

\begin{table}[ht]
\centering
\subfloat[][]{\begin{tabular}{|c|c|c|c|} \hline
Feature vector & $\begin{array}{ccc}1 & \simboloX & \simboloX \end{array}$ & $\begin{array}{ccc}\simboloX & 1 & \simboloX \end{array}$ & $\begin{array}{ccc}\simboloX & \simboloX & 1\end{array}$ \\ \hline\hline
Empirical Probability & $0.0748$ & $0.0330$ & $0.0227$ \\ \hline
\end{tabular}\label{tab:indepFeat20Q_a}} \\
\subfloat[][]{\begin{tabular}{|c|c|c|c|} \hline
Feature vector & $\begin{array}{ccc}1 & 1 & \simboloX\end{array}$ & $\begin{array}{ccc}1 & \simboloX & 1\end{array}$ & $\begin{array}{ccc}\simboloX & 1 & 1\end{array}$ \\ \hline\hline
Empirical Probability & $0.0028$ & $0.0012$ & $0.0009$ \\ \hline
Product Probability & $0.0025$ & $0.0017$ & $0.0007$ \\ \hline
\end{tabular}\label{tab:indepFeat20Q_b}} \\
\caption{Probabilities of possessing at least (a) one latent feature, or (b) two latent features, as given in the patterns shown in the heading rows. The symbol `$\simboloX$' denotes either $0$ or $1$. The `empirical probability' rows contain the probabilities extracted directly from the inferred IBP matrix $\Zb$, while the `product probability' row shows the product of the corresponding two latent feature probabilities given in (a).}
\label{tab:indepFeat20Q}
\end{table}



\subsection{Comorbidity analysis of personality disorders}

In order to identify the seven personality disorders studied in the previous section, psychiatrists have established specific diagnostic criteria for each of them. These criteria correspond to affirmative responses to one or several questions in the NESARC survey and this correspondence is shown in Appendix~\ref{app:Crit-Quest}. Then, there exists a set of criteria to identify if a subject presents any of the following personality disorders: avoidant, dependent, obsessive-compulsive, paranoid, schizoid, histrionic and antisocial. In the present analysis, we consider as input data the fulfillment of the $52$ criteria (i.e., $R=2$) corresponding to all the disorders for the  $43{,}093$ subjects and we apply the variational inference algorithm truncated to $K=25$ features, as detailed in Section~\ref{sec:variational}, to find the latent structure of the data. 

In order to properly initialize the huge amount of variational parameters, we have previously run six Gibbs samplers over the data but taking only the criteria corresponding to the avoidant PD and another PD (i.e., the seven criteria for the avoidant PD and the seven for the dependent PD, the criteria for the avoidant PD with the eight for the OCPD, etc.) for $10,000$ randomly chosen subjects. After running the six Gibbs samplers, we obtain $18$ latent features that we group in a unique matrix $\Zb$ to obtain the weighting matrices $\Bb^d_\MAP$, which are used to initialize some parameters $\nu_{nk}$ and $\phi^d_{kr}$. We do this because the variational algorithm is sensitive to the starting point and a random initialization would not produce good solutions.

\begin{figure}[ht]
\centering
\includegraphics[width=\textwidth]{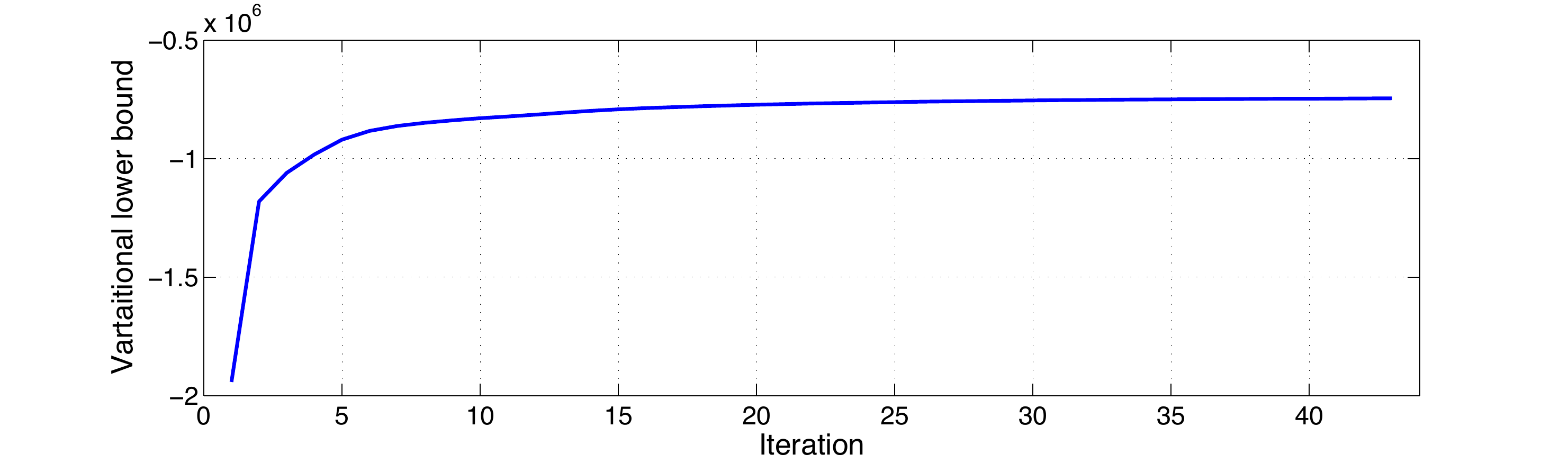}
\caption{Variational lower bound $ \mathcal{L}(\Hcal, \Hcal_q)$ at each iteration.}
\label{fig:cotaIt}
\end{figure}

We run enough iterations of the variational algorithm to ensure convergence of the variational lower bound (the lower bound at each iteration is shown in Figure~\ref{fig:cotaIt}).  We construct a binary matrix $\Zb$ by setting each element $z_{nk}=1$ if $\nu_{nk}>0.5$. We flip (changing zeros by ones, and vice versa) those features possessed by more than $80\%$ of the subjects, obtaining only $10$ latent features possessed by more than $50$ subjects among the $43,093$ in the database and then recomputing the weighting matrices. In Table~\ref{Tab:pU_52Q}, we show the probability of occurrence of each feature (top row), as well as the probability of having active only one single feature (bottom row). We also show  the `empirical' and the `product' probabilities of possessing at least two latent features in Table~\ref{Tab: prob conjunta}, and the probabilities of possessing at least two features given that one of them is active in Table~\ref{Tab:prob condicional}.

\begin{figure}[H]
\centering
\includegraphics[width=1\textwidth, angle=0]{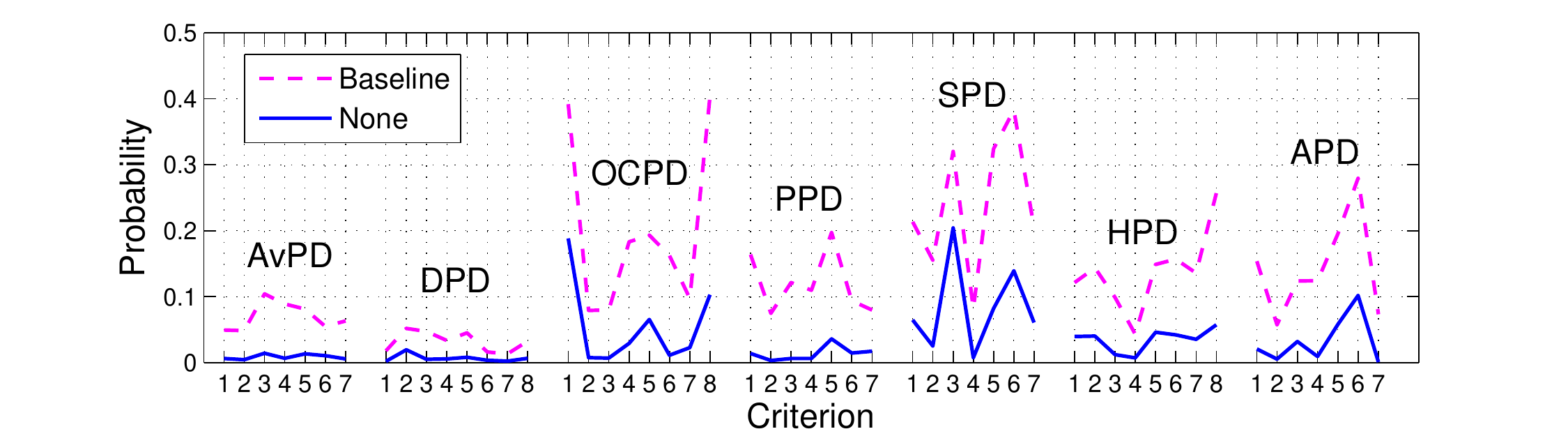}
\caption{Probability of meeting each criterion. The probabilities when no latent feature is active (solid curve) have been obtained using the matrices $\Bb^d_{\MAP}$, while the baseline (dashed curve) has been obtained taking into account the $43,093$ subjects in the database. \newline (AvPD=Avoidant PD, DPD=Dependent PD, OCPD=Obsessive-compulsive PD, PPD=Paranoid PD, SPD=Schizoid PD, HPD=Histrionic PD, APD=Antisocial PD)}
\label{fig:52Q_1}
\end{figure}

In Figure \ref{fig:52Q_1}, we plot the probability of meeting each criterion in the general population (dashed line) and the probability of meeting each criterion for those subjects that do not have any active feature in our model (solid line). There are $15,185$ subjects (35.2\% of the population) which do not present any active feature, and for these people the probability of meeting any criterion is reduced significantly. 

\begin{figure}[H]
\centering
\includegraphics[width=1\textwidth, angle=0]{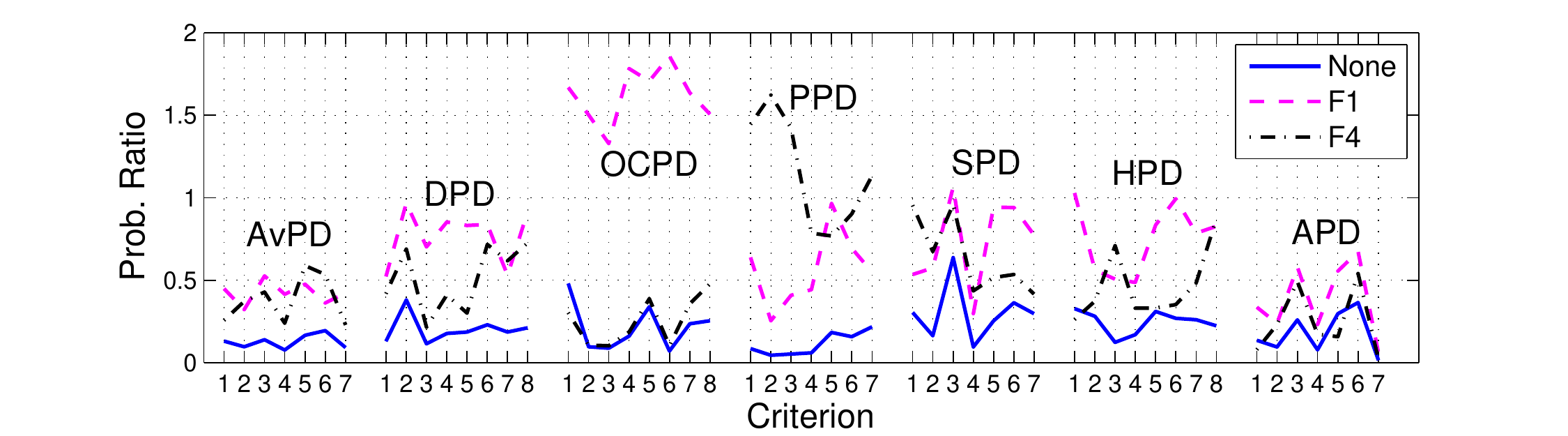}\\
\includegraphics[width=1\textwidth, angle=0]{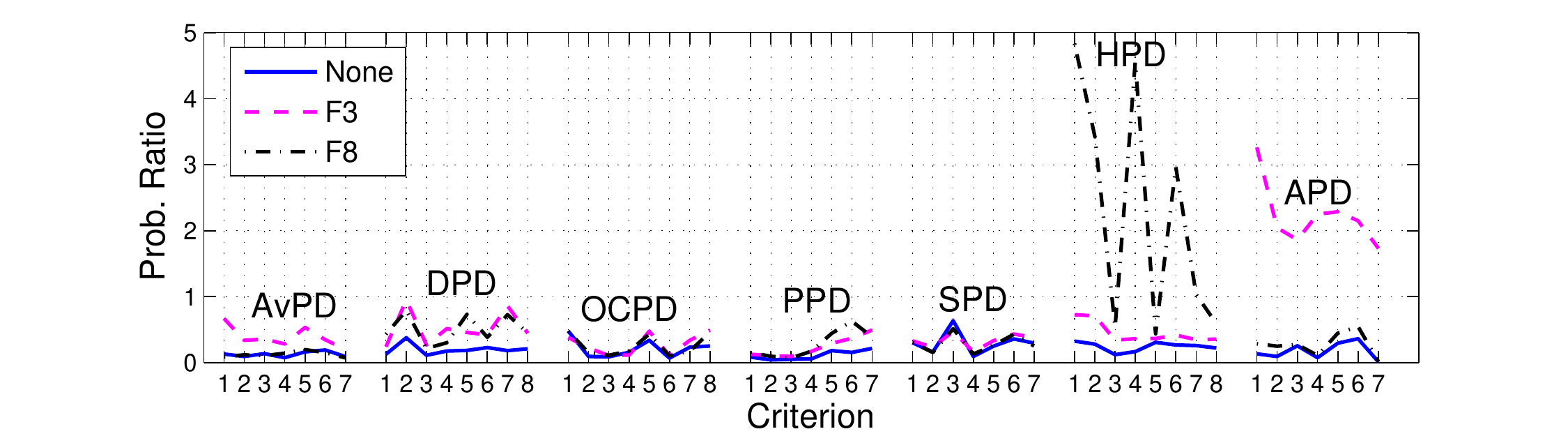}\\
\includegraphics[width=1\textwidth, angle=0]{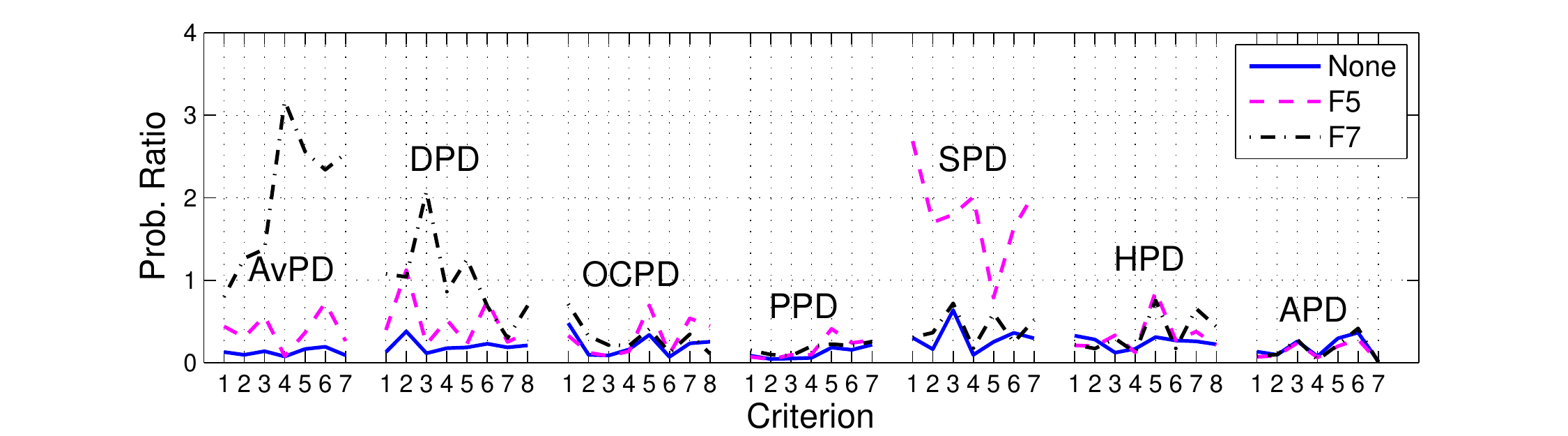}
\caption{Probability ratio of meeting each criterion, with respect to the baseline. These probabilities have been obtained using the matrices $\Bb^d_{\MAP}$, when none or a single feature is active (the legend shows the active latent features). }
\label{fig:52Q_2}
\end{figure}

We have found results that are in accordance with previous studies and at the same time provide new information to understand personality disorders. Out of the 10 features, 6 of them directly describe personality disorders. Feature 1 increases the probability of fulfilling the criteria for OCPD, Feature 3 increases the probability of fulfilling the criteria for antisocial, Feature 4 increases the probability of fulfilling the criteria for paranoid, Feature 5 increases the probability of meeting the criteria for schizoid, Feature 8 increases the probability of fulfilling the criteria for histrionic and Feature 7 increases the probability of meeting the criteria for avoidant and dependent. In Figure \ref{fig:52Q_2}, we plot the probability ratio between the probability of meeting each criterion when a single feature is active with respect to the probability of meeting each criterion in the general population (baseline in Figure~\ref{fig:52Q_1}). So, if the ratio is above one, it means that the feature increases the probability of meeting that criterion with respect to the general population. In all these plots, we also show the probability ratio between not having any active feature and the general population, which serves as a reference for a low probability of fulfilling a criterion. Note that the scale on the vertical axis may be different through all the figures for a better display. In Figure~\ref{fig:52Q_2}, we can see that only the criteria for one of the personality disorders is systematically above one, when one feature is active, except for Feature 7 that increases the probability for both avoidant and dependent. In the figure, we can also notice that when one feature is active the probability of the criteria for the other disorders is above the probability for the subjects that do not have any active feature, although lower than the general population (above the solid line and below one). It partially shows the comorbidity pattern for each personality disorder. For example, Feature 1, besides increasing the probability of meeting the criteria for OCPD, also increases the probability of meeting criterion 3 for schizoid and criterion 1 for histrionic. It is also important to point out that Feature 8 increases significantly the probability of meeting criteria 1, 2, 4 and 6 for histrionic (and mildly for criterion 7), but it does not affect criteria 3, 5 and 8, although the probability of meeting these criteria are increased by Feature 4 (paranoid) and Feature 5 (schizoid). In a way, it indicates that criteria 3 and 8 are more related to paranoid disorder and criterion 5 to schizoid disorder.

\begin{figure}[H]
\centering
\includegraphics[width=1\textwidth, angle=0]{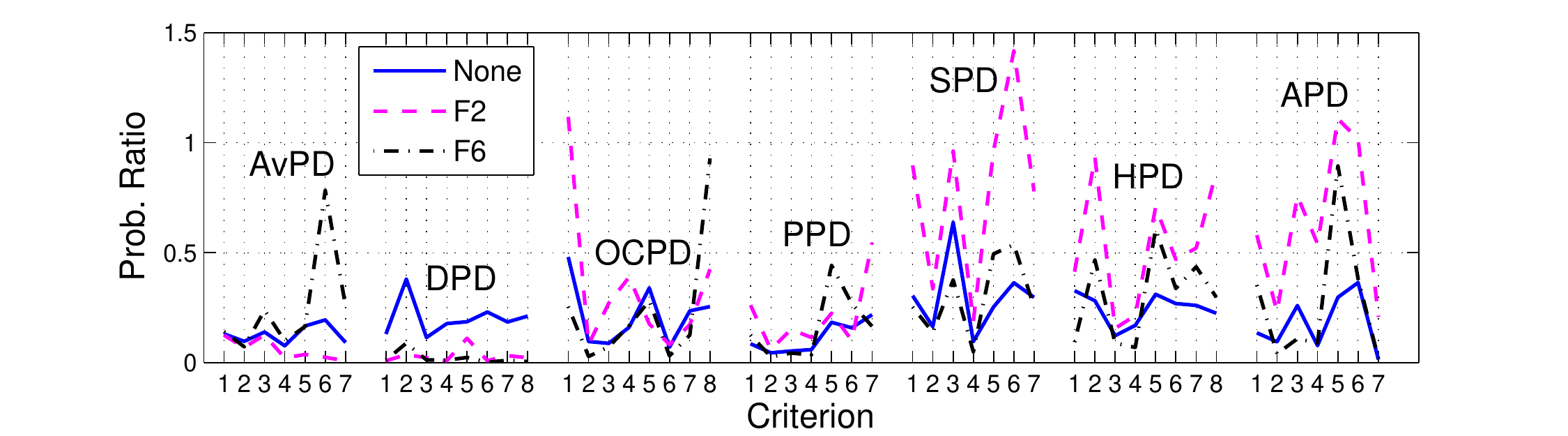}\\
\caption{Probability ratio of meeting each criterion, with respect to the baseline. These probabilities have been obtained using the matrices $\Bb^d_{\MAP}$, when none or a single feature is active (the legend shows the active latent features). }
\label{fig:52Q_3}
\end{figure}

\begin{figure}[H]
\centering
\includegraphics[width=1\textwidth, angle=0]{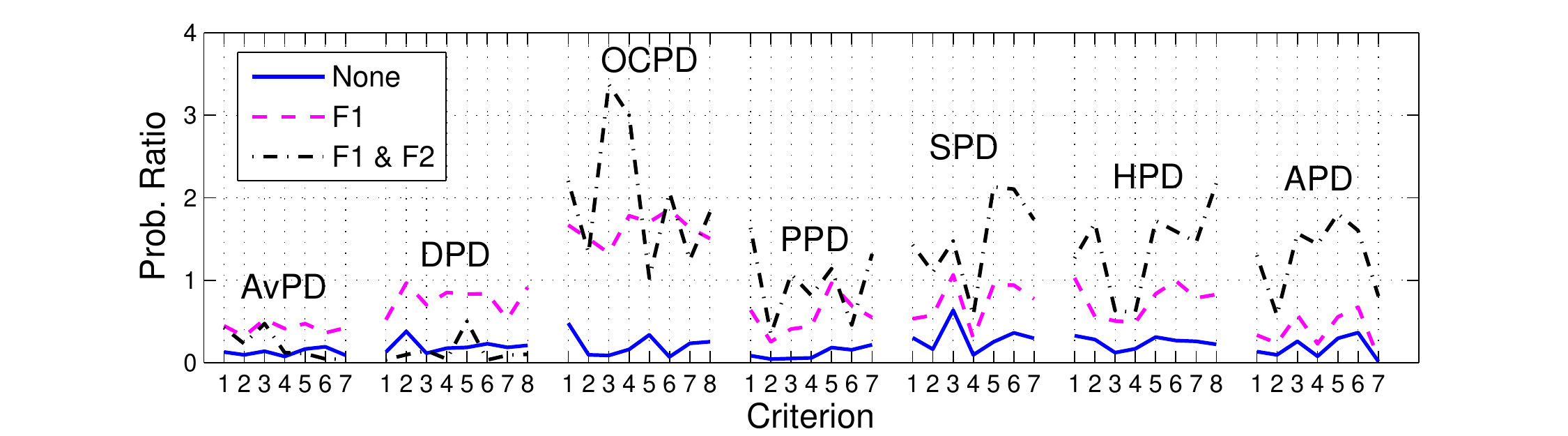}\\
\includegraphics[width=1\textwidth, angle=0]{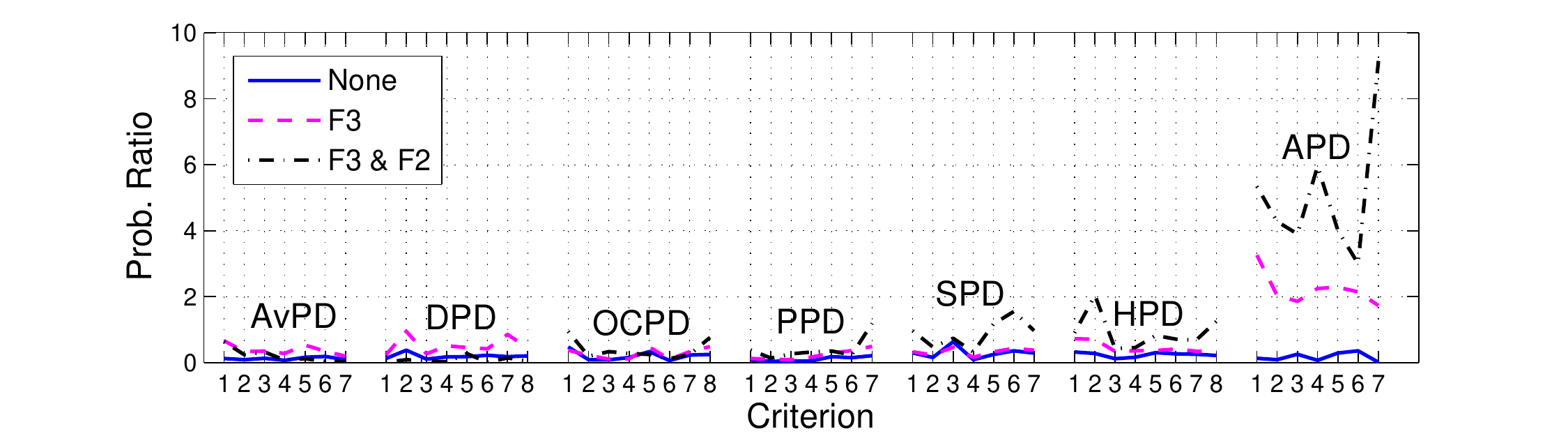}\\
\includegraphics[width=1\textwidth, angle=0]{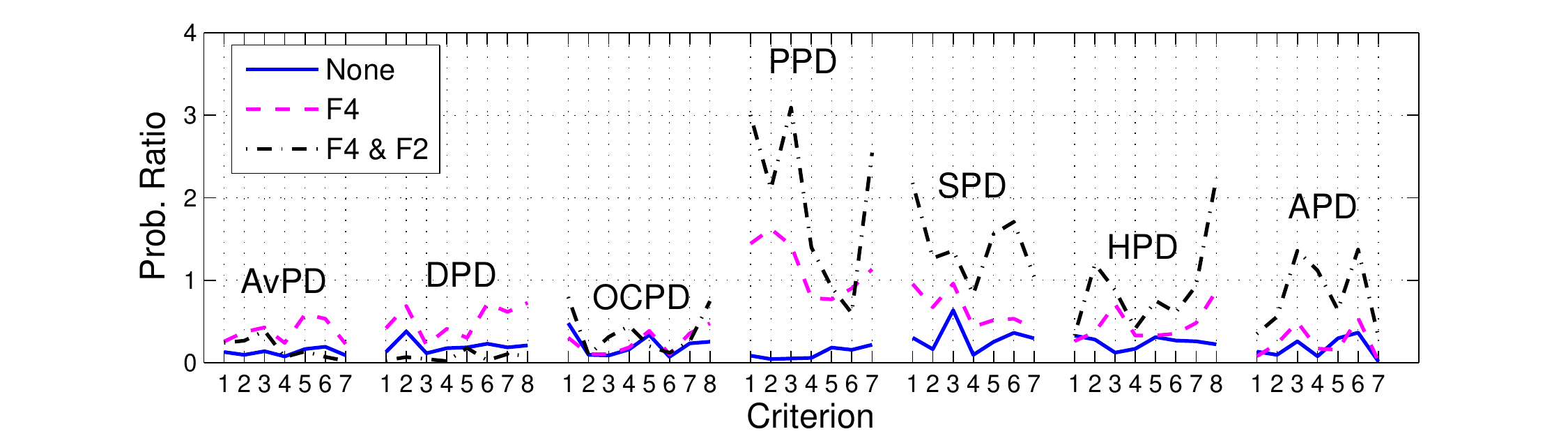}\\
\includegraphics[width=1\textwidth, angle=0]{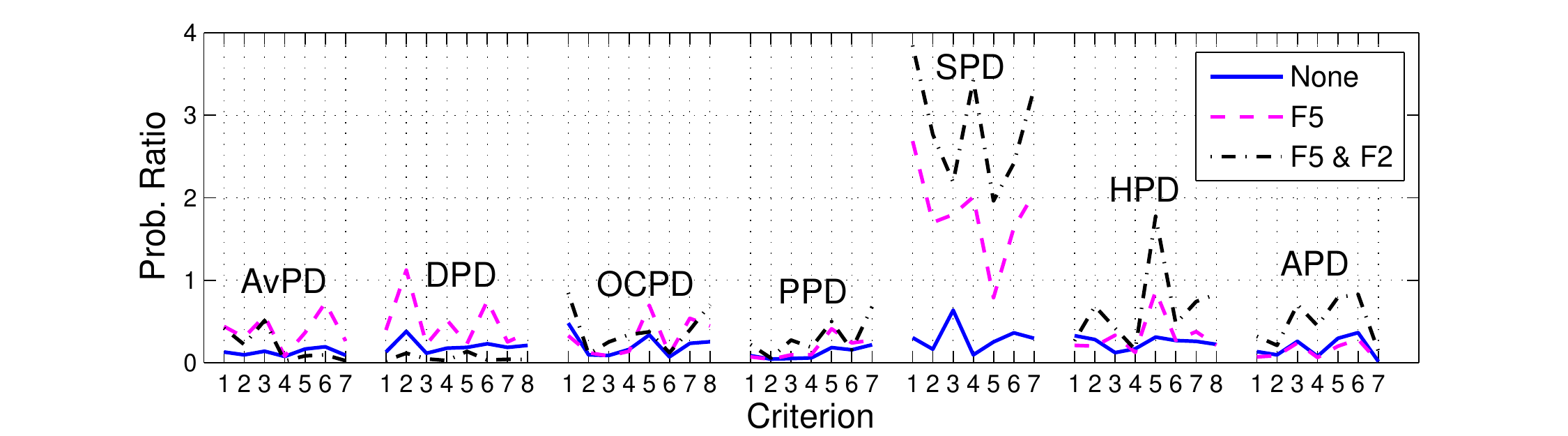}
\caption{Probability ratio of meeting each criterion, with respect to the baseline. These probabilities have been obtained using the matrices $\Bb^d_{\MAP}$, when none, a single or two features are active (the legend shows the active latent features). }
\label{fig:52Q_4}
\end{figure}

As seen in Figure \ref{fig:52Q_3}, Features 2 and 6 mainly reduce the probability of meeting the criteria for dependent PD. Feature 2 also reduces criteria 4-7 for avoidant and mildly increases criterion 1 for OCPD, criterion 6 for schizoid and criteria 5 and 6 for antisocial. Feature 6 also reduces some criteria below the probability for the subjects with no active features. But for most of the criteria the probability ratio moves between one and the ratio for the subjects with no active feature. When these features appear by themselves, the subjects might be similar to the subjects without any active feature, they become relevant when they appear together with other features. These features are less likely to be isolated features than the previous ones, as reported in Table~\ref{Tab:pU_52Q}. For example, Feature 2 appears frequently with Features 1, 3, 4 and 5, as shown in Table~\ref{Tab:prob condicional}, and the probability ratios are plotted in Figure \ref{fig:52Q_4} and compared to the probability ratio when each feature is not accompanied by Feature 2. We can see that when we add Feature 2 to Feature 1, the comorbidity pattern changes significantly and it results in subjects with higher probabilities of meeting the criteria for every other disorder except avoidant and dependent. Additionally, when we add Feature 2 to Feature 5, we can see that meeting the criteria for schizoid is even more probable, together with criterion 5 for histrionic. 

\begin{figure}[H]
\centering
\includegraphics[width=1\textwidth, angle=0]{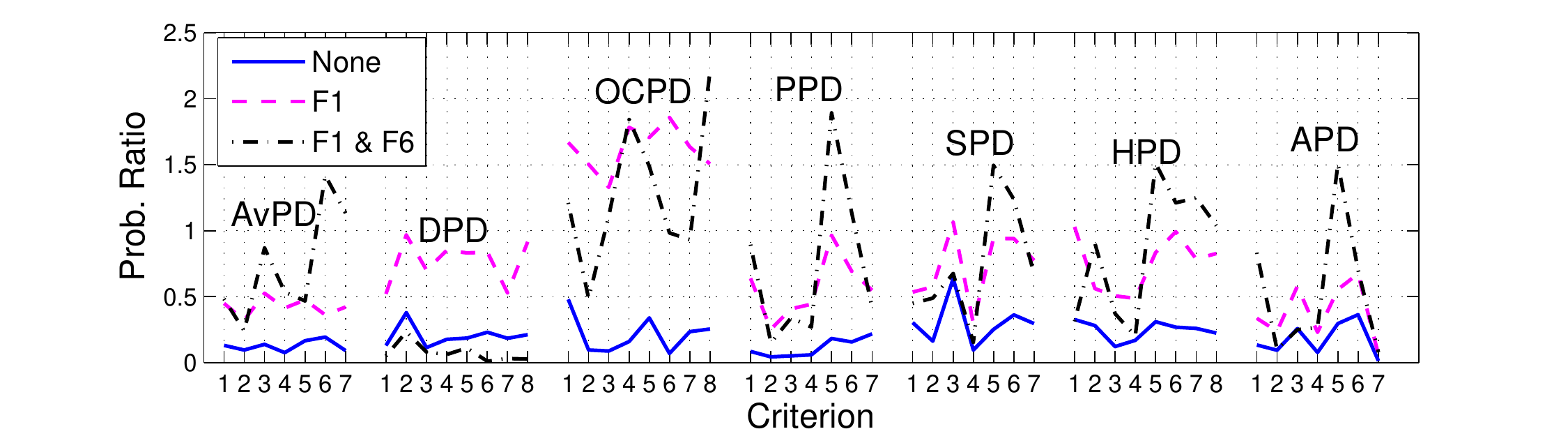}\\
\includegraphics[width=1\textwidth, angle=0]{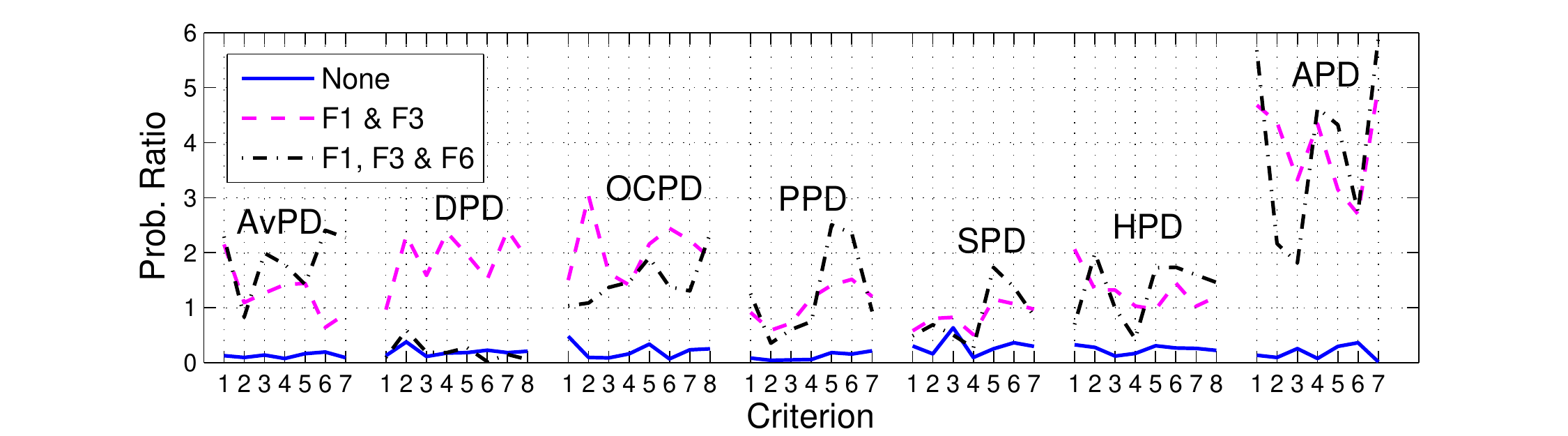}
\caption{Probability ratio of meeting each criterion, with respect to the baseline. These probabilities have been obtained using the matrices $\Bb^d_{\MAP}$, when none, a single or several features are active (the legend shows the active latent features). }
\label{fig:52Q_5}
\end{figure}

Either Feature 1 or Features 1 and 3 typically accompany Feature 6, and Feature 6 is seldom seen by itself (see Tables~\ref{Tab:pU_52Q} and \ref{Tab:Lista20}).  In Figure \ref{fig:52Q_5}, we show the probability ratio when Feature 1 is active and when Features 1 and 3 are active, as reference, and when we add Feature 6 to them. Adding Feature 6 mainly reduces the probability of meeting the criteria for dependent. It is also relevant to point out that Features 1 and 3 increase the probability of meeting the criteria 5 and 6 for paranoid, while Feature 4 mainly increased the probability of meeting the criteria 1-4 for paranoid personality disorder, as shown in Figure~\ref{fig:52Q_2}.

Feature 9 is similar to Feature 7, as it captures an increase in the probability of meeting the criteria for avoidant and dependent, but it never appears isolated and most times it appears together with Features 1 and 4.

\begin{figure}[H]
\centering
\includegraphics[width=1\textwidth, angle=0]{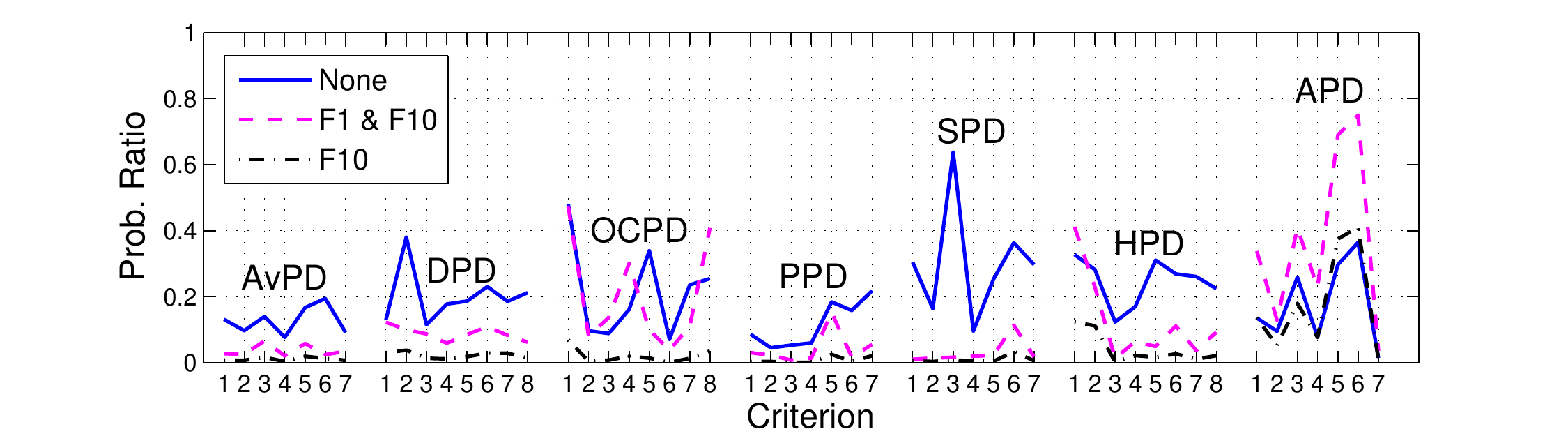}
\caption{Probability ratio of meeting each criterion, with respect to the baseline. These probabilities have been obtained using the matrices $\Bb^d_{\MAP}$, when none, a single  or two features are active (the legend shows the active latent features).}
\label{fig:52Q_6}
\end{figure}

Feature 10 never appears isolated and it mainly appears only with Feature 1. This feature by itself only indicates that the probability of all the criteria should be much lower than the subjects with no active features, except for antisocial, which behaves as the subjects with no active features. When we add Feature 1 to Feature 10, we get that the probability of meeting the criteria for OCDP goes to that of the subject with no active features, as can be seen in Figure \ref{fig:52Q_6}. For us this is a spurious feature that is equivalent to not having any active feature and that the variational algorithm has not been able to eliminate. This is always a risk when working with flexible models, like BNP, in which a spurious component might appear when it should not. These components can be eliminated by common sense in most cases or by further analysis by experts (psychiatric experts in our case). But it can also indicate an unknown component that can point towards a new research direction previously unknown, which is one of the attractive features of using generative models.

\begin{figure}[H]
\centering
\includegraphics[width=1\textwidth, angle=0]{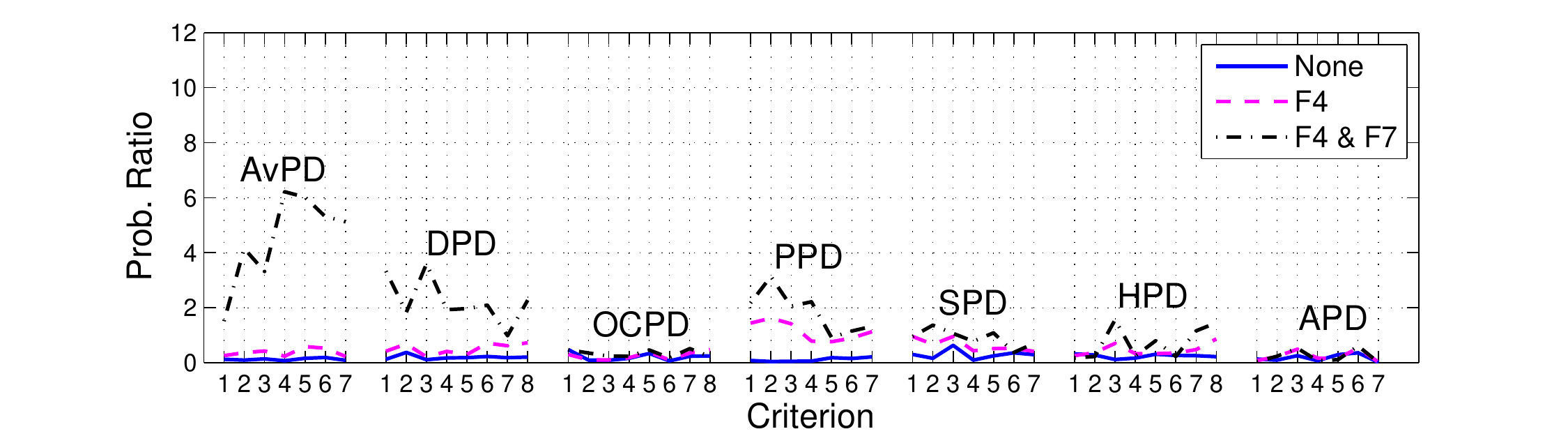}\\
\includegraphics[width=1\textwidth, angle=0]{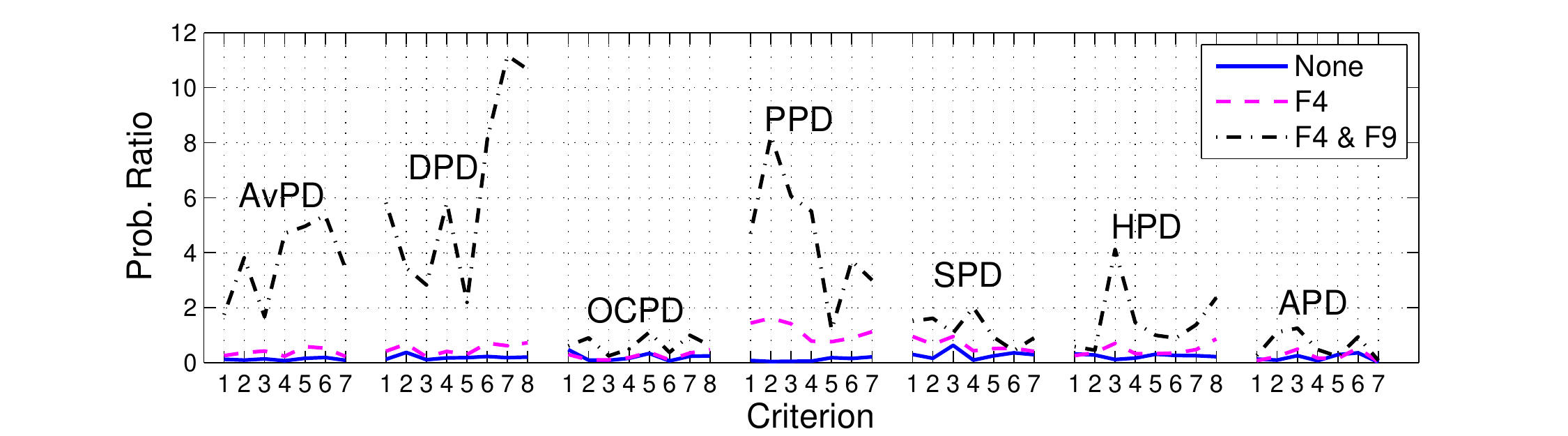}
\caption{Probability ratio of meeting each criterion, with respect to the baseline. These probabilities have been obtained using the matrices $\Bb^d_{\MAP}$, when none, a single  or two features are active (the legend shows the active latent features).}
\label{fig:52Q_7}
\end{figure}

Besides the comorbidity patterns shown by the individual features that we have already reported, we can also see that almost all the features are positively correlated. In Table~\ref{Tab: prob conjunta}, we show the probability that any two features appear together (upper triangular sub-matrix) and the joint probability that we should observe if the features were independent (lower triangular sub-matrix). Ignoring Feature 10, all of the other features are positively correlated, except Features 2 and 7 and Features 8 and 5 that seem uncorrelated (the differences are not statistically significant). Most of the features are strongly correlated and the differences in Table~\ref{Tab: prob conjunta} correspond to several standard deviations higher (between 3 and 42) that we should expect from independent random observations. For example, the correlation between Features 4 and 9 and Features 4 and 7 is quite high and both show subjects with higher probability of meeting the criteria for avoidant, dependent and paranoid. The difference between Features 7 and 9 is given by the criteria 1-4 for paranoid PD, that are significantly increased by Feature 9 and slightly by Feature 7, as it can be seen in Figure \ref{fig:52Q_7}. Finally, it is worth mentioning that Feature 4 (paranoid) is the most highly correlated feature with all the others, so we can say that anyone suffering from paranoid PD has a higher comorbidity with any other personality disorder. 

\section{Conclusions}\label{sec:conclusions}
In this paper, we have proposed a new model that combines the IBP with discrete observations using the multinomial-logit distribution. We have used the Laplace approximation to integrate out the weighting factors, which allows us to efficiently run the Gibbs sampler. We have also derived a variational inference algorithm, which allows dealing with larger databases and provides accurate results.

We have applied our model to the NESARC database to find out the hidden features that characterize the psychiatric disorders. First, we have used the Gibbs sampler to extract the latent structure behind $20$ of the most common psychiatric disorders. As a result, we have found that the comorbidity patterns of these psychiatric disorders can be described by only three latent features, which mainly model the internalizing disorders, the externalizing disorders, and a general psychopathology factor. Additionally, we have applied the variational inference algorithm to analyze the relation among the $52$ criteria defined by the psychiatrists to diagnose each of the seven personality disorders (i.e., externalizing disorders). We have obtained that for most of the disorders, a latent feature appears to model all the criteria that characterize that particular disorder. In this experiment, we have also seen that avoidant and dependent PDs are jointly modeled by four features, and that paranoid disorder is the most highly correlated PD with all the others.


\acks{Francisco J. R. Ruiz is supported by an FPU fellowship from the Spanish Ministry of Education (AP2010-5333), Isabel Valera is supported by  the \textit{Plan Regional-Programas I+D} of \textit{Comunidad de Madrid} (AGES-CM S2010/BMD-2422), Fernando P\'erez-Cruz has been partially supported by a Salvador de Madariaga grant, and Carlos Blanco acknowledges NIH grants (DA019606 and DA023200) and the New York State Psychiatric Institute for their support.
The authors also acknowledge the support of \textit{Ministerio de Ciencia e Innovaci\'{o}n} of Spain (projects DEIPRO TEC2009-14504-C02-00, ALCIT TEC2012-38800-C03-01, and program Consolider-Ingenio 2010 CSD2008-00010 COMONSENS).
This work was also supported by the European Union 7th Framework Programme through the Marie Curie Initial Training Network ``Machine Learning for Personalized Medicine'' MLPM2012, Grant No. 316861.
}


\begin{table}[H]
\centering
\begin{tabular}{|c|c|c|c|c|c|c|c|c|c|c|}
\hline
{Features} & $1$ & $2$ &$3$ &$4$& $5$ & $6$ &$7$ &$8$ &$9$ & $10$\\
\hline \hline
Total & $43.45$ &   $19.01$ &   $15.28$  & $13.99$ &  $11.76$  &  $8.97$ &  $ 7.54$  &  $6.91$  &  $1.86$  &  $1.43$\\
\hline
Single feature & $13.48$  &  $3.62$  &  $2.22$  &  $1.34$  &  $2.27$  &  $0.49$ &  $ 0.76$  &  $1.07$   &      $0 $   &     $0$ \\
\hline
\end{tabular}
\caption{ Probabilities ($\%$) of possessing (top row) at least one latent feature, or (bottom row) a single feature.}\label{Tab:pU_52Q}
\end{table}

\begin{table}[H]
\centering
\begin{tabular}{|c|c|c|c|c|c|c|c|c|c|c|}
\hline
Features & $1$ & $2$ &$3$ &$4$& $5$ & $6$ &$7$ &$8$ &$9$ & $10$\\
 \hline \hline 
$1$ & \cellcolor[rgb]{0.9,0.5,0.1} & $9.92$ & $8.96$ & $8.48$ & $5.67$ & $7.22$ & $4.92$ & $3.85$ & $1.46$ & $1.42$ \\ \hline
$2$ & \cellcolor[cmyk]{0,0,0.4,0} $8.26$ & \cellcolor[rgb]{0.9,0.5,0.1} & $4.43$ & $4.54$ & $3.67$ & $1.90$ & $1.43$ & $2.08$ & $0.71$ & $0.21$ \\ \hline
$3$ & \cellcolor[cmyk]{0,0,0.4,0} $6.64$ & \cellcolor[cmyk]{0,0,0.4,0} $2.90$ & \cellcolor[rgb]{0.9,0.5,0.1} & $3.29$ & $2.18$ & $3.00$ & $2.02$ & $1.58$ & $0.54$ & $0.20$ \\ \hline
$4$ & \cellcolor[cmyk]{0,0,0.4,0} $6.08$ & \cellcolor[cmyk]{0,0,0.4,0} $2.66$ & \cellcolor[cmyk]{0,0,0.4,0} $2.14$ & \cellcolor[rgb]{0.9,0.5,0.1} & $2.79$ & $1.91$ & $2.39$ & $1.40$ & $1.25$ & $0.03$ \\ \hline
$5$ & \cellcolor[cmyk]{0,0,0.4,0} $5.11$ & \cellcolor[cmyk]{0,0,0.4,0} $2.23$ & \cellcolor[cmyk]{0,0,0.4,0} $1.80$ & \cellcolor[cmyk]{0,0,0.4,0} $1.65$ & \cellcolor[rgb]{0.9,0.5,0.1} & $1.31$ & $1.35$ & $0.85$ & $0.57$ & $0.00$ \\ \hline
$6$ & \cellcolor[cmyk]{0,0,0.4,0} $3.90$ & \cellcolor[cmyk]{0,0,0.4,0} $1.71$ & \cellcolor[cmyk]{0,0,0.4,0} $1.37$ & \cellcolor[cmyk]{0,0,0.4,0} $1.26$ & \cellcolor[cmyk]{0,0,0.4,0} $1.05$ & \cellcolor[rgb]{0.9,0.5,0.1} & $1.10$ & $0.80$ & $0.44$ & $0.14$ \\ \hline
$7$ & \cellcolor[cmyk]{0,0,0.4,0} $3.28$ & \cellcolor[cmyk]{0,0,0.4,0} $1.43$ & \cellcolor[cmyk]{0,0,0.4,0} $1.15$ & \cellcolor[cmyk]{0,0,0.4,0} $1.06$ & \cellcolor[cmyk]{0,0,0.4,0} $0.89$ & \cellcolor[cmyk]{0,0,0.4,0} $0.68$ & \cellcolor[rgb]{0.9,0.5,0.1} & $0.65$ & $0.28$ & $0.00$ \\ \hline
$8$ & \cellcolor[cmyk]{0,0,0.4,0} $3.00$ & \cellcolor[cmyk]{0,0,0.4,0} $1.31$ & \cellcolor[cmyk]{0,0,0.4,0} $1.06$ & \cellcolor[cmyk]{0,0,0.4,0} $0.97$ & \cellcolor[cmyk]{0,0,0.4,0} $0.81$ & \cellcolor[cmyk]{0,0,0.4,0} $0.62$ & \cellcolor[cmyk]{0,0,0.4,0} $0.52$ & \cellcolor[rgb]{0.9,0.5,0.1} & $0.51$ & $0.07$ \\ \hline
$9$ & \cellcolor[cmyk]{0,0,0.4,0} $0.81$ & \cellcolor[cmyk]{0,0,0.4,0} $0.35$ & \cellcolor[cmyk]{0,0,0.4,0} $0.28$ & \cellcolor[cmyk]{0,0,0.4,0} $0.26$ & \cellcolor[cmyk]{0,0,0.4,0} $0.22$ & \cellcolor[cmyk]{0,0,0.4,0} $0.17$ & \cellcolor[cmyk]{0,0,0.4,0} $0.14$ & \cellcolor[cmyk]{0,0,0.4,0} $0.13$ & \cellcolor[rgb]{0.9,0.5,0.1} & $0.00$ \\ \hline
$10$ & \cellcolor[cmyk]{0,0,0.4,0} $0.62$ & \cellcolor[cmyk]{0,0,0.4,0} $0.27$ & \cellcolor[cmyk]{0,0,0.4,0} $0.22$ & \cellcolor[cmyk]{0,0,0.4,0} $0.20$ & \cellcolor[cmyk]{0,0,0.4,0} $0.17$ & \cellcolor[cmyk]{0,0,0.4,0} $0.13$ & \cellcolor[cmyk]{0,0,0.4,0} $0.11$ & \cellcolor[cmyk]{0,0,0.4,0} $0.10$ & \cellcolor[cmyk]{0,0,0.4,0} $0.03$ & \cellcolor[rgb]{0.9,0.5,0.1} \\ \hline
\end{tabular}
\caption{Probabilities ($\%$) of possessing at least two latent features. The elements above the diagonal correspond to the `empirical probability', i.e., extracted directly from the inferred IBP matrix $\Zb$, and the elements below the diagonal correspond to the `product probability' of the corresponding two latent feature probabilities given in the first row of Table \ref{Tab:pU_52Q}.}\label{Tab: prob conjunta}
\end{table}

\begin{table}[H]
\centering
\begin{tabular}{|c|c|c|c|c|c|c|c|c|c|c|}
\hline
\backslashbox{$k_1$}{$k_2$} & $1$ & $2$ &$3$ &$4$& $5$ & $6$ &$7$ &$8$ &$9$ & $10$\\
 \hline \hline 
$1$ & $100$ & $22.83$ & $20.63$ & $19.53$ & $13.05$ & $16.62$ & $11.33$ & $8.85$ & $3.37$ & $3.27$ \\ \hline
$2$ & $52.19$ & $100$ & $23.33$ & $23.90$ & $19.32$ & $10.00$ & $7.51$ & $10.95$ & $3.75$ & $1.09$ \\ \hline
$3$ & $58.68$ & $29.03$ & $100$ & $21.54$ & $14.29$ & $19.66$ & $13.25$ & $10.34$ & $3.51$ & $1.29$ \\ \hline
$4$ & $60.63$ & $32.47$ & $23.52$ & $100$ & $19.97$ & $13.65$ & $17.05$ & $10.02$ & $8.92$ & $0.20$ \\ \hline
$5$ & $48.22$ & $31.25$ & $18.57$ & $23.77$ & $100$ & $11.11$ & $11.49$ & $7.24$ & $4.88$ & $0.00$ \\ \hline
$6$ & $80.47$ & $21.18$ & $33.47$ & $21.29$ & $14.56$ & $100$ & $12.23$ & $8.92$ & $4.86$ & $1.53$ \\ \hline
$7$ & $65.26$ & $18.92$ & $26.83$ & $31.63$ & $17.91$ & $14.55$ & $100$ & $8.65$ & $3.66$ & $0.03$ \\ \hline
$8$ & $55.62$ & $30.11$ & $22.86$ & $20.28$ & $12.32$ & $11.58$ & $9.43$ & $100$ & $7.39$ & $1.07$ \\ \hline
$9$ & $78.46$ & $38.23$ & $28.77$ & $67.00$ & $30.76$ & $23.41$ & $14.82$ & $27.40$ & $100$ & $0.12$ \\ \hline
$10$ & $99.19$ & $14.40$ & $13.75$ & $1.94$ & $0.00$ & $9.55$ & $0.16$ & $5.18$ & $0.16$ & $100$ \\ \hline
\end{tabular}
\caption{Probabilities  ($\%$) of possessing at least features $k_1$ and $k_2$ given that $k_1$ is active, i.e., $\left(\sum_{n=1}^N z_{n k_1} z_{n k_2} \right)/\left(\sum_{n=1}^N z_{n k_1}\right)$.}\label{Tab:prob condicional}
\end{table}

\begin{table}[H]
\centering
\begin{tabular}{|c|c|c|c|c|c|c|c|c|c|c|c|}
\cline{2-12}
\multicolumn{1}{c|}{} & \multirow{2}{*}{\textbf{\# Occurrences}} & \multicolumn{10}{c|}{\textbf{Features}} \\ \cline{3-12}
\multicolumn{1}{c|}{} & & $1$ & $2$ & $3$ & $4$ & $5$ & $6$ & $7$ & $8$ & $9$ & $10$ \\ \hline
$1$ & $15185$ & $0$ & $0$ & $0$ & $0$ & $0$ & $0$ & $0$ & $0$ & $0$ & $0$ \\ \hline
$2$ & $5811$ & $1$ & $0$ & $0$ & $0$ & $0$ & $0$ & $0$ & $0$ & $0$ & $0$ \\ \hline
$3$ & $1561$ & $0$ & $1$ & $0$ & $0$ & $0$ & $0$ & $0$ & $0$ & $0$ & $0$ \\ \hline
$4$ & $1389$ & $1$ & $1$ & $0$ & $0$ & $0$ & $0$ & $0$ & $0$ & $0$ & $0$ \\ \hline
$5$ & $1021$ & $1$ & $0$ & $1$ & $0$ & $0$ & $0$ & $0$ & $0$ & $0$ & $0$ \\ \hline
$6$ & $977$ & $0$ & $0$ & $0$ & $0$ & $1$ & $0$ & $0$ & $0$ & $0$ & $0$ \\ \hline
$7$ & $958$ & $1$ & $0$ & $0$ & $0$ & $0$ & $1$ & $0$ & $0$ & $0$ & $0$ \\ \hline
$8$ & $956$ & $0$ & $0$ & $1$ & $0$ & $0$ & $0$ & $0$ & $0$ & $0$ & $0$ \\ \hline
$9$ & $946$ & $1$ & $0$ & $0$ & $1$ & $0$ & $0$ & $0$ & $0$ & $0$ & $0$ \\ \hline
$10$ & $687$ & $1$ & $0$ & $0$ & $0$ & $1$ & $0$ & $0$ & $0$ & $0$ & $0$ \\ \hline
$11$ & $576$ & $0$ & $0$ & $0$ & $1$ & $0$ & $0$ & $0$ & $0$ & $0$ & $0$ \\ \hline
$12$ & $553$ & $1$ & $0$ & $0$ & $0$ & $0$ & $0$ & $1$ & $0$ & $0$ & $0$ \\ \hline
$13$ & $495$ & $0$ & $1$ & $0$ & $0$ & $1$ & $0$ & $0$ & $0$ & $0$ & $0$ \\ \hline
$14$ & $486$ & $1$ & $0$ & $0$ & $0$ & $0$ & $0$ & $0$ & $1$ & $0$ & $0$ \\ \hline
$15$ & $460$ & $0$ & $0$ & $0$ & $0$ & $0$ & $0$ & $0$ & $1$ & $0$ & $0$ \\ \hline
$16$ & $451$ & $0$ & $1$ & $1$ & $0$ & $0$ & $0$ & $0$ & $0$ & $0$ & $0$ \\ \hline
$17$ & $438$ & $1$ & $0$ & $0$ & $0$ & $0$ & $0$ & $0$ & $0$ & $0$ & $1$ \\ \hline
$18$ & $414$ & $1$ & $0$ & $1$ & $0$ & $0$ & $1$ & $0$ & $0$ & $0$ & $0$ \\ \hline
$19$ & $385$ & $0$ & $1$ & $0$ & $1$ & $0$ & $0$ & $0$ & $0$ & $0$ & $0$ \\ \hline
$20$ & $370$ & $1$ & $1$ & $0$ & $1$ & $0$ & $0$ & $0$ & $0$ & $0$ & $0$ \\ \hline
\end{tabular}
\caption{List of the $20$ most common feature patterns.}\label{Tab:Lista20}
\end{table}


\appendix

\section{Laplace approximation details}\label{app:Laplace}
In this section we provide the necessary details for the implementation of the Laplace approximation proposed in Section~\ref{sec:laplace}. The expression in \eqref{eq:PsiDef} can be rewritten as
\begin{equation}\label{eq:PsiDef2}
\begin{split}
 f(\Bb^d) & = \trace\left\{{\Mb^d}^\top\Bb^d\right\}-\sumd_{n=1}^N \log\left( \sumd_{r=1}^R\exp (\zn \br^d) \right) \\
 & \quad -\frac{1}{2\s2B}\trace\left\{{\Bb^d}^\top\Bb^d\right\}-\frac{R(K+1)}{2}\log(2\pi\s2B),
\end{split}
\end{equation}
where $(\Mb^d)_{kr}$ counts the number of data points for which $x_{nd}=r$ and $z_{nk}=1$, namely,
$(\Mb^d)_{kr}=\sum_{n=1}^N\delta(x_{nd}=r)z_{nk}$,
where $\delta(\cdot)$ is the Kronecker delta function. By definition, $(\Mb^d)_{0r}=\sum_{n=1}^N\delta(x_{nd}=r)$.


By defining $(\rd)_{kr} = \sumd_{n=1}^{N} z_{nk}\pi_{nd}^r$, the gradient of $f(\Bb^d)$ can be derived as 
\begin{equation}\label{gradiente}
 \nabla f = \Mb^d-\rd-\frac{1}{\s2B}\Bb^d.
\end{equation}

To compute the Hessian, it is easier to define the gradient $ \nabla f $ as a vector, instead of a matrix, and hence we stack the columns of $\Bb^d$ into $\bd$, i.e., $\bd=\Bb^d(:)$ for avid Matlab users. The Hessian matrix can now be readily computed taking the derivatives of the gradient, yielding
\begin{align}
  \nabla\nabla f &= -\frac{1}{\s2B}\Ib_{R(K+1)}+\nabla\nabla\log p(\xd | \bd,\Zb)\nonumber\\&= -\frac{1}{\s2B}\Ib_{R(K+1)}-\sumd_{n=1}^N \left(\diag(\pind)-(\pind)^\top \pind\right)\otimes (\zn^\top \zn),
\end{align}
where $\diag(\pind)$ is a diagonal matrix with the values of the vector $\pind= \left[\pi^1_{nd}, \pi^2_{nd}, \ldots,\pi^R_{nd}\right]$ as its diagonal elements.

Finally, note that, since $p(\xd | \bd,\Zb)$ is a log-concave function of $\bd$ \cite[p.~87]{Boyd}, $-\nabla\nabla f$ is a positive definite matrix, which guarantees that the maximum of $f(\bd)$ is unique.

\section{Lower Bound derivation}\label{app:lowerBound}
In this section we derive the lower bound $\mathcal{L}(\Hcal,\Hcal_q)$ on the evidence $p(\Xb|\Hcal)$. From Eq.~\ref{eq:defCotaVariac},
\begin{equation}\label{eqA:defCotaVariac}
\begin{split}
 \log p(\Xb|\Hcal) & = \mathbb{E}_q \left[ \log p(\Psi,\Xb|\Hcal) \right]  + H[q] +D_{KL}(q||p) \\
 & \geqslant \mathbb{E}_q \left[ \log p(\Psi,\Xb|\Hcal) \right]  + H[q].
\end{split}
\end{equation}

The expectation $\mathbb{E}_q \left[ \log  p(\Psi,\Xb|\Hcal) \right]$ can be derived as
\begin{equation}\label{eq:E(log p)}
\begin{split}
\mathbb{E}_q \left[ \log  p(\Psi,\Xb|\Hcal) \right]&=  \sumd_{k=1}^{K} \underbrace{\mathbb{E}_q \left[\log p(v_k|\alpha)  \right]}_{1}+  \sumd_{d=1}^{D}\sumd_{k=1}^{K}  \underbrace{\mathbb{E}_q \left[\log p(\bk^d | \s2B) \right]}_{2} + \sumd_{d=1}^{D} \underbrace{\mathbb{E}_q \left[\log p(\bb_0^d | \s2B) \right]}_3\\
&  \quad + \sumd_{k=1}^{K}\sumd_{n=1}^{N} \underbrace{\mathbb{E}_q \left[\log  p(z_{nk}|\{v_i\}_{i=1}^{k}) \right]}_4 +\sumd_{n=1}^{N}\sumd_{d=1}^{D} \underbrace{\mathbb{E}_q \left[\log p(x_{nd} | \zn, \Bb^d,\bb_0^d) \right]}_5
,
\end{split}
\end{equation}
where each term can be computed as shown below:

\begin{enumerate}
\item For the Beta distribution over $v_k$,
\begin{equation}
\mathbb{E}_q \left[\log p(v_k|\alpha) \right]= \log (\alpha)+ (\alpha -1) \left[\psi(\tau_{k1})- \psi(\tau_{k1}+ \tau_{k2}) \right].
\end{equation}
 
 \item  For the Gaussian distribution over vectors $\bk^d$,
\begin{equation}
\mathbb{E}_q \left[\log p(\bk^d | \s2B) \right]= -\frac{R}{2} \log (2\pi \s2B) - \frac{1}{2\s2B} \left( \sumd_{r=1}^{R}(\phi^d_{kr})^2+ \sumd_{r=1}^{R} \skrd  \right).
\end{equation} 

\item   For the Gaussian distribution over $\bb_0^d $,
\begin{equation}
\mathbb{E}_q \left[\log p(\bb_0^d | \s2B) \right]= -\frac{R}{2} \log (2\pi \s2B) - \frac{1}{2\s2B} \left( \sumd_{r=1}^{R}(\phi^d_{0r})^2+ \sumd_{r=1}^{R} (\sigma^d_{0r})^2  \right).
\end{equation} 

\item For the feature assignments, which are Bernoulli distributed given the feature probabilities, we have
\begin{equation}
\begin{split}
\mathbb{E}_q \left[\log p(z_{nk}|\{v_i\}_{i=1}^{k})   \right] = & (1- \nu_{nk}) \mathbb{E}_q \left[\log \left(1- \prodd_{i=1}^k v_i \right)\right]\\
& + \nu_{nk} \sumd_{i=1}^{k} \left[\psi(\tau_{i1})- \psi(\tau_{i1}+ \tau_{i2}) \right],
\end{split}
\end{equation}
where the expectation $\mathbb{E}_q \left[\log \left(1- \prod_{i=1}^k v_i \right)\right] $ has no closed-form solution. We can instead lower bound it by using the multinomial approach \cite[]{DoshiVelez}. Under this approach, we introduce an auxiliary multinomial distribution $\boldsymbol{\lambda}_k= [\lambda_{k1},\ldots,\lambda_{kk}]$ in the expectation and apply Jensen's inequality, yielding
\begin{equation}
\begin{split}
 \mathbb{E}_q \left[\log \left(1- \prodd_{i=1}^k v_i \right)\right] \geq & \sumd_{m=1}^{k} \lambda_{km} \psi( \tau_{m2}) + \sumd_{m=1}^{k-1} \left( \sumd_{n=m+1}^{k} \lambda_{kn} \right) \psi( \tau_{m1})\\
 & - \sumd_{m=1}^{k} \left( \sumd_{n=m}^{k} \lambda_{kn} \right) \psi( \tau_{m1} + \tau_{m2}) - \sumd_{m=1}^{k} \lambda_{km} \log (\lambda_{km}),
 \end{split}
\end{equation}
which holds for any distribution represented by the probabilities $\lambda_{k1},\ldots,\lambda_{kk}$, for $1\leq k\leq K$. Then,
\begin{equation}
\begin{split}
\mathbb{E}_q \left[\log p(z_{nk}|\{v_i\}_{i=1}^{k})   \right] 
& \geq (1- \nu_{nk}) \left[\sumd_{m=1}^{k} \lambda_{km} \psi( \tau_{m2}) + \sumd_{m=1}^{k-1} \left( \sumd_{n=m+1}^{k} \lambda_{kn} \right) \psi( \tau_{m1}) \right.\\
 & \left. \quad - \sumd_{m=1}^{k} \left( \sumd_{n=m}^{k} \lambda_{kn} \right) \psi( \tau_{m1} + \tau_{m2}) - \sumd_{m=1}^{k} \lambda_{km} \log (\lambda_{km})\right] \\
 & \quad + \nu_{nk} \sumd_{i=1}^{k} \left[\psi(\tau_{i1})- \psi(\tau_{i1}+ \tau_{i2}) \right].
\end{split}
\end{equation}

\item For the likelihood term, we can write
%
%
\begin{equation}
\mathbb{E}_q \left[\log p(x_{nd} | \zn, \Bb^d, \bb_0^d) \right]= \phi^d_{0x_{nd}}+\sumd_{k=1}^{K} \nu_{nk} \phi^d_{kx_{nd}} - \mathbb{E}_q \left[\log \left( \sumd_{r=1}^{R}\exp(\zn \br^d + \b0r^d)  \right) \right],
\end{equation}
where the logarithm can be upper bounded by its first-order Taylor series expansion around the auxiliary variable $\xi_{nd}^{-1}$ (for $n=1,\ldots,N$ and $d=1,\ldots,D$) \cite[]{Blei,Bouchard}, yielding
\begin{equation}
\begin{split}
&\log  \left( \sumd_{r=1}^{R}\exp(\zn \br^d + \b0r^d)  \right) \leq \xi_{nd} \left( \sumd_{r=1}^{R}\exp(\zn \br^d + \b0r^d)  \right) - \log(\xi_{nd})-1.
\end{split}
\end{equation}
The main advantage of this bound lies on the fact that it allows us to compute the expectation of the bound for the Gaussian distribution, since it involves the moment generating functions of the distributions $q(\br^d)$ and $q(\b0r^d)$. Then, we can lower bound the likelihood term as
\begin{equation}
\begin{split}
&\mathbb{E}_q \left[\log p(x_{nd} | \zn, \Bb^d, \bb_0^d) \right] \geq \phi^d_{0x_{nd}}+\sumd_{k=1}^{K} \nu_{nk} \phi^d_{kx_{nd}} + \log(\xi_{nd})+1\\
&\quad  - \xi_{nd} \sumd_{r=1}^{R} \left[ \exp\left(\phi^d_{0r}+\frac{1}{2} (\sigma^{d}_{0r})^2 \right) \prodd_{k=1}^{K} \left( 1-\nu_{nk}+ \nu_{nk} \exp\left(\phi^d_{kr}+\frac{1}{2} \skrd \right)\right)  \right].\end{split}
\end{equation}
\end{enumerate}

Substituting the previous results in \eqref{eq:E(log p)}, we obtain 
\begin{equation}
\begin{split}
\mathbb{E}_q \left[ \log  p(\Psi,\Xb|\Hcal) \right]&\geq  \sumd_{k=1}^{K}  \left[\log (\alpha)+ (\alpha -1) \left(\psi(\tau_{k1})- \psi(\tau_{k1}+ \tau_{k2}) \right) \right]\\
& \quad -\frac{R(K+1)D}{2} \log (2\pi \s2B) - \frac{1}{2\s2B} \sumd_{k=0}^{K}\sumd_{d=1}^{D} \sumd_{r=1}^{R}\left((\phi^d_{kr})^2+ \skrd  \right)\\
&\quad +\sumd_{n=1}^{N} \sumd_{k=1}^{K}\left[\nu_{nk} \sumd_{i=1}^{k} \left[\psi(\tau_{i1})- \psi(\tau_{i1}+ \tau_{i2}) \right]\right. \\
& \left. \quad \quad\quad \quad  \quad \quad+ (1- \nu_{nk}) \left(\sumd_{m=1}^{k} \lambda_{km} \psi( \tau_{m2}) + \sumd_{m=1}^{k-1} \left( \sumd_{n=m+1}^{k} \lambda_{kn} \right) \psi( \tau_{m1}) \right. \right. \\
& \left. \left. \quad\quad \quad \quad  \quad \quad - \sumd_{m=1}^{k} \left( \sumd_{n=m}^{k} \lambda_{kn} \right) \psi( \tau_{m1} + \tau_{m2}) - \sumd_{m=1}^{k} \lambda_{km} \log (\lambda_{km})\right) \right]\\
& \quad +\sumd_{n=1}^{N}\sumd_{d=1}^{D} \left[ \phi^d_{0x_{nd}}+\sumd_{k=1}^{K} \nu_{nk} \phi^d_{kx_{nd}} + \log(\xi_{nd})+1 \right.\\
&\quad \left.  - \xi_{nd} \sumd_{r=1}^{R} \left[ \exp\left(\phi^d_{0r}+\frac{1}{2} (\sigma^{d}_{0r})^2 \right) \prodd_{k=1}^{K} \left( 1-\nu_{nk}+ \nu_{nk} \exp\left(\phi^d_{kr}+\frac{1}{2} \skrd \right)\right)  \right]\right].
\end{split}
\end{equation}

Additionally, the entropy of the distribution $q(\Psi)$ is given by
\begin{equation}
\begin{split}
H[q]&= \mathbb{E}_q \left[\log q(\Psi)\right]\\
&= \sumd_{k=1}^{K } \mathbb{E}_q \left[\log q(v_k|\tau_{k1},\tau_{k2})\right] +\sumd_{d=1}^{D } \sumd_{r=1}^{R } \sumd_{k=0}^{K } \mathbb{E}_q \left[\log q(b^d_{kr}|\phi_{kr}^d,\skrd)\right]  + \sumd_{n=1}^{N } \sumd_{k=1}^{K } \mathbb{E}_q \left[\log q(z_{nk}|\nu_{nk})\right] \\
&= \sumd_{k=1}^{K } \left[\log\left(\frac{\Gamma(\tau_{k1}) \Gamma(\tau_{k2})}{\Gamma(\tau_{k1}+ \tau_{k2})}\right) - (\tau_{k1}-1) \psi(\tau_{k1}) - (\tau_{k2}-1) \psi(\tau_{k2})  +(\tau_{k1}+\tau_{k2}-2) \psi(\tau_{k1} +\tau_{k2})\right] \\
& \quad +\sumd_{d=1}^{D } \sumd_{r=1}^{R } \sumd_{k=0}^{K } \frac{1}{2} \log(2\pi e \skrd)+ \sumd_{n=1}^{N } \sumd_{k=1}^{K } \left[ -\nu_{nk} \log(\nu_{nk}) - (1-\nu_{nk}) \log (1-\nu_{nk})\right]. 
\end{split}
\end{equation}

Finally, we obtain the lower bound on the evidence $p(\Xb|\Hcal)$ as
\begin{equation}\label{eq:defCotaVariacFinal}
\begin{split}
 &\log p(\Xb|\Hcal)  \geq \mathbb{E}_q \left[ \log p(\Psi,\Xb|\Hcal) \right]  + H[q]\\
 &\geq \sumd_{k=1}^{K}  \left[\log (\alpha)+ (\alpha -1) \left(\psi(\tau_{k1})- \psi(\tau_{k1}+ \tau_{k2}) \right) \right]\\
& \quad -\frac{R(K+1)D}{2} \log (2\pi \s2B) - \frac{1}{2\s2B} \sumd_{k=0}^{K}\sumd_{d=1}^{D} \sumd_{r=1}^{R}\left((\phi^d_{kr})^2+ \skrd  \right)\\
&\quad +\sumd_{n=1}^{N} \sumd_{k=1}^{K}\left[\nu_{nk} \sumd_{i=1}^{k} \left[\psi(\tau_{i1})- \psi(\tau_{i1}+ \tau_{i2}) \right]\right. \\
& \left. \quad \quad\quad \quad  \quad \quad+ (1- \nu_{nk}) \left(\sumd_{m=1}^{k} \lambda_{km} \psi( \tau_{m2}) + \sumd_{m=1}^{k-1} \left( \sumd_{n=m+1}^{k} \lambda_{kn} \right) \psi( \tau_{m1}) \right. \right. \\
& \left. \left. \quad\quad \quad \quad  \quad \quad - \sumd_{m=1}^{k} \left( \sumd_{n=m}^{k} \lambda_{kn} \right) \psi( \tau_{m1} + \tau_{m2}) - \sumd_{m=1}^{k} \lambda_{km} \log (\lambda_{km})\right) \right]\\
& \quad +\sumd_{n=1}^{N}\sumd_{d=1}^{D} \left[ \phi^d_{0x_{nd}}+\sumd_{k=1}^{K} \nu_{nk} \phi^d_{kx_{nd}} + \log(\xi_{nd})+1 \right.\\
&\quad \left.  - \xi_{nd} \sumd_{r=1}^{R} \left[ \exp\left(\phi^d_{0r}+\frac{1}{2} (\sigma^{d}_{0r})^2 \right) \prodd_{k=1}^{K} \left( 1-\nu_{nk}+ \nu_{nk} \exp\left(\phi^d_{kr}+\frac{1}{2} \skrd \right)\right)  \right]\right]\\
& \quad +\sumd_{k=1}^{K } \left[\log\left(\frac{\Gamma(\tau_{k1}) \Gamma(\tau_{k2})}{\Gamma(\tau_{k1}+ \tau_{k2})}\right) - (\tau_{k1}-1) \psi(\tau_{k1}) - (\tau_{k2}-1) \psi(\tau_{k2})  +(\tau_{k1}+\tau_{k2}-2) \psi(\tau_{k1} +\tau_{k2})\right] \\
& \quad +\sumd_{d=1}^{D } \sumd_{r=1}^{R } \sumd_{k=0}^{K } \frac{1}{2} \log(2\pi e \skrd)+ \sumd_{n=1}^{N } \sumd_{k=1}^{K } \left[ -\nu_{nk} \log(\nu_{nk}) - (1-\nu_{nk}) \log (1-\nu_{nk})\right] \\
&=  \mathcal{L}(\Hcal, \Hcal_q),
\end{split}
\end{equation}
where $\Hcal_q= \{ \tau_{k1},\tau_{k2}, \lambda_{km}, \xi_{nd}, \nu_{nk} , \phi^d_{kr}, \phi^d_{0r}, \skrd, (\sigma^d_{0r})^2\} $ (for $k=1,\ldots,K$, $m=1,\ldots,k$, $d=1,\ldots,D$, and $n=1,\ldots,N$) represents the set of the variational parameters. 

\section{Derivatives for Newton's method}\label{app:derivatives}
- For the parameters of the Gaussian distribution  $q(b^d_{kr} | \phi^d_{kr},\skrd) $ for $k= 1,\ldots,K$,
\begin{equation}\label{eq:1st_deriv}
\begin{split}
\frac{\partial}{\partial \phi^d_{kr}} \mathcal{L}(\Hcal, \Hcal_q) &=  - \frac{1}{\s2B} \phi^d_{kr} +  \sumd_{n=1}^{N} \Bigg[ \nu_{nk} \delta(x_{nd}=r)- \nu_{nk}\xi_{nd}  \exp\left(\phi^d_{0r}+\frac{1}{2} (\sigma^{d}_{0r})^2 \right)\exp\left(\phi^d_{kr}+\frac{1}{2} \skrd \right)    \\
& \quad \quad \quad \quad \quad \quad \quad \quad \quad   \times \prodd_{k'\neq k} \left( 1-\nu_{nk'}+ \nu_{nk'} \exp\left(\phi^d_{k'r}+\frac{1}{2} (\sigma^d_{k'r})^2 \right)\right)  \Bigg].
\end{split}
\end{equation}

\begin{equation}
\begin{split}
\frac{\partial^2}{\partial (\phi^d_{kr})^2} \mathcal{L}(\Hcal, \Hcal_q) &=  - \frac{1}{\s2B}  -  \sumd_{n=1}^{N} \Bigg[ \nu_{nk} \xi_{nd}  \exp\left(\phi^d_{0r}+\frac{1}{2} (\sigma^{d}_{0r})^2 \right)\exp\left(\phi^d_{kr}+\frac{1}{2} \skrd \right)    \\
& \quad \quad \quad \quad \quad \quad \quad   \times \prodd_{k'\neq k} \left( 1-\nu_{nk'}+ \nu_{nk'} \exp\left(\phi^d_{k'r}+\frac{1}{2} (\sigma^d_{k'r})^2 \right)\right)  \Bigg].
\end{split}
\end{equation}

\begin{equation}
\begin{split}
\frac{\partial}{\partial \skrd} \mathcal{L}(\Hcal, \Hcal_q) &=  - \frac{1}{2\s2B}+\frac{1}{2} (\sigma^d_{kr})^{-2} - \frac{1}{2}\sumd_{n=1}^{N} \Bigg[   \nu_{nk} \xi_{nd}  \exp\left(\phi^d_{0r}+\frac{1}{2} (\sigma^{d}_{0r})^2 \right)\exp\left(\phi^d_{kr}+\frac{1}{2} \skrd \right)    \\
& \quad  \times \prodd_{k'\neq k} \left( 1-\nu_{nk'}+ \nu_{nk'} \exp\left(\phi^d_{k'r}+\frac{1}{2} (\sigma^d_{k'r})^2 \right)\right)  \Bigg].
\end{split}
\end{equation}

\begin{equation}
\begin{split}
\frac{\partial^2}{(\partial \skrd)^2} \mathcal{L}(\Hcal, \Hcal_q) &=  -\frac{1}{2} (\sigma^d_{kr})^{-4} - \frac{1}{4}\sumd_{n=1}^{N} \Bigg[   \nu_{nk} \xi_{nd}  \exp\left(\phi^d_{0r}+\frac{1}{2} (\sigma^{d}_{0r})^2 \right)\exp\left(\phi^d_{kr}+\frac{1}{2} \skrd \right)    \\
& \quad  \times \prodd_{k'\neq k} \left( 1-\nu_{nk'}+ \nu_{nk'} \exp\left(\phi^d_{k'r}+\frac{1}{2} (\sigma^d_{k'r})^2 \right)\right)  \Bigg].
\end{split}
\end{equation}

- For the parameters of the Gaussian distribution $ q(b^d_{0r} | \phi^d_{0r}, (\sigma^d_{0r})^2)$,
\begin{equation}
\begin{split}
&\frac{\partial}{\partial \phi^d_{0r}} \mathcal{L}(\Hcal, \Hcal_q) \\
 &=  - \frac{1}{\s2B}\phi^d_{0r} + \sumd_{n=1}^{N} \Bigg[ \delta(x_{nd}=r) -\xi_{nd}  \exp\left(\phi^d_{0r}+\frac{1}{2} (\sigma^{d}_{0r})^2 \right) \prodd_{k=1}^{K} \left( 1-\nu_{nk}+ \nu_{nk} \exp\left(\phi^d_{kr}+\frac{1}{2} \skrd \right)\right)  \Bigg].
\end{split}
\end{equation}

\begin{equation}
\begin{split}
&\frac{\partial^2}{(\partial \phi^d_{0r} )^2} \mathcal{L}(\Hcal, \Hcal_q) \\
&=   - \frac{1}{\s2B}- \sumd_{n=1}^{N} \Bigg[  \xi_{nd}  \exp\left(\phi^d_{0r}+\frac{1}{2} (\sigma^{d}_{0r})^2 \right) \prodd_{k=1}^{K} \left( 1-\nu_{nk}+ \nu_{nk} \exp\left(\phi^d_{kr}+\frac{1}{2} \skrd \right)\right)  \Bigg].
\end{split}
\end{equation}

\begin{equation}
\begin{split}
&\frac{\partial}{\partial (\sigma^d_{0r})^2} \mathcal{L}(\Hcal, \Hcal_q)   \\
&=  - \frac{1}{2\s2B}+ \frac{1}{2}(\sigma^d_{0r})^{-2}-\frac{1}{2} \sumd_{n=1}^{N} \Bigg[  \xi_{nd}  \exp\left(\phi^d_{0r}+\frac{1}{2} (\sigma^{d}_{0r})^2 \right) \prodd_{k=1}^{K} \left( 1-\nu_{nk}+ \nu_{nk} \exp\left(\phi^d_{kr}+\frac{1}{2} \skrd \right)\right)  \Bigg].
\end{split}
\end{equation}

\begin{equation}\label{eq:last_deriv}
\begin{split}
&\frac{\partial^2}{(\partial (\sigma^d_{0r})^2 )^2} \mathcal{L}(\Hcal, \Hcal_q) \\
&=   -\frac{1}{2}(\sigma^d_{0r})^{-4} -\frac{1}{4} \sumd_{n=1}^{N} \Bigg[  \xi_{nd}  \exp\left(\phi^d_{0r}+\frac{1}{2} (\sigma^{d}_{0r})^2 \right) \prodd_{k=1}^{K} \left( 1-\nu_{nk}+ \nu_{nk} \exp\left(\phi^d_{kr}+\frac{1}{2} \skrd \right)\right)  \Bigg].
\end{split}
\end{equation}

\newpage
\section{Correspondence between criteria and questions in NESARC}\label{app:Crit-Quest}
\begin{table}[H]
\small
\begin{tabular}{|c|c|}
\hline
\textbf{Question Code} & \textbf{Personality disorder and criterion}\\
\hline \hline
S10Q1A1-S10Q1B7 & Avoidant (1 question for each diagnostic criterion)\\\hline\hline
S10Q1A8-S10Q1B15 & Dependent (1 question for each diagnostic criterion)\\\hline\hline
S10Q1A16-S10Q1B17 & OCPD criterion 1\\
S10Q1A18-S10Q1B23 & OCPD criteria 2-7\\
S10Q1A24-S10Q1B25 & OCPD criterion 8\\\hline\hline
S10Q1A26-S10Q1B29 & Paranoid criteria 1-4\\
S10Q1A30-S10Q1A31 & Paranoid criterion 5\\
S10Q1A32-S10Q1B33 & Paranoid criteria 6-7\\\hline\hline
S10Q1A45-S10Q1B46 & Schizoid criterion 1\\
S10Q1A47-S10Q1B48 & Schizoid criteria 2-3\\
S10Q1A50-S10Q1B50 & Schizoid criterion 4\\
S10Q1A43-S10Q1B43 & Schizoid criterion 5\\
S10Q1A51-S10Q1B52 & Schizoid criterion 6\\
S10Q1A49-S10Q1B49 or S10Q1A53-S10Q1B53 & Schizoid criterion 7\\\hline \hline
S10Q1A54-S10Q1B54 or S10Q1A56-S10Q1B56  & Histrionic criterion 1\\
S10Q1A58-S10Q1B58 or S10Q1A60-S10Q1B60  & Histrionic criterion 2\\
S10Q1A55-S10Q1B55  & Histrionic criterion 3\\
S10Q1A61-S10Q1B61 & Histrionic criterion 4\\
S10Q1A64-S10Q1B64 & Histrionic criterion 5\\
S10Q1A59-S10Q1B59 or S10Q1A62-S10Q1B62  & Histrionic criterion 6\\
S10Q1A63-S10Q1B63  & Histrionic criterion 7\\
S10Q1A57-S10Q1B57   & Histrionic criterion 8\\\hline \hline
S11Q1A20-S11Q1A25 & Antisocial, criterion 1\\
S11Q1A11- S11Q1A13 &  Antisocial, criterion 2\\
S11Q1A8- S11Q1A10 & Antisocial, criterion 3\\
S11Q1A17- S11Q1A18 & Antisocial, criterion 4\\
S11Q1A26- S11Q1A33 & Antisocial, criterion 4\\
S11Q1A14- S11Q1A16 & Antisocial, criterion 5\\
S11Q1A6 and S11Q1A19 & Antisocial, criterion 6\\
S11Q8A-B & Antisocial, criterion 7\\
\hline
\end{tabular}
\normalsize
\caption{Correspondence between the criteria for each personality disorder and questions in NESARC.}
\end{table}


%
%

\vskip 0.2in
\bibliography{Bib}

\end{document}